\documentclass[preprint,12pt,authoryear]{elsarticle}

\usepackage{amssymb}
\usepackage{graphics} 
\usepackage{epsfig} 
\usepackage{mathptmx} 
\usepackage{times} 
\usepackage{amsmath} 
\usepackage{adjustbox}
\usepackage{subcaption}
\usepackage{bm} 
\usepackage{booktabs}
\usepackage{url}
\usepackage[T1]{fontenc}
\usepackage{graphicx}
\usepackage{adjustbox}
\usepackage[capitalise]{cleveref}
\usepackage{multirow,tabularx}
\usepackage[table, svgnames]{xcolor}
\journal{Journal of \LaTeX}

\begin{document}

\begin{frontmatter}



\title{Out-of-distribution detection in 3D applications: A review}


\author[label1]{Zizhao Li}
\ead{zizhao.li1@student.unimelb.edu.au}
\author[label1,label2]{Xueyang Kang\corref{cor1}}
\ead{xueyang.kang@student.unimelb.edu.au;kangxueyang@126.com}
\author[label1]{Joseph West}
\ead{joseph.west@unimelb.edu.au}
\author[label1]{Kourosh Khoshelham} 
\ead{k.khoshelham@unimelb.edu.au}
\affiliation[label1]{organization={The University of Melbourne},
             city={Parkville},
             postcode={3010}, 
             state={Victoria},
             country={Australia}}

\affiliation[label2]{organization={KU Leuven},
             city={Leuven},
             postcode={3000}, 
             state={Flemish Brabant},
             country={Belgium}}

\cortext[cor1]{Corresponding author. E-mail address: kangxueyang@126.com (Xueyang Kang).}

\begin{abstract}

The ability to detect objects that are not prevalent in the training set is a critical capability in many 3D applications, including autonomous driving. Machine learning methods for object recognition often assume that all object categories encountered during inference belong to a closed set of classes present in the training data. This assumption limits generalization to the real world, as objects not seen during training may be misclassified or entirely ignored. As part of reliable AI, OOD detection identifies inputs that deviate significantly from the training distribution. This paper provides a comprehensive overview of OOD detection within the broader scope of trustworthy and uncertain AI. We begin with key use cases across diverse domains, introduce benchmark datasets spanning multiple modalities, and discuss evaluation metrics. Next, we present a comparative analysis of OOD detection methodologies, exploring model structures, uncertainty indicators, and distributional distance taxonomies, alongside uncertainty calibration techniques. Finally, we highlight promising research directions, including adversarially robust OOD detection and failure identification, particularly relevant to 3D applications. The paper offers both theoretical and practical insights into OOD detection, showcasing emerging research opportunities such as 3D vision integration. These insights help new researchers navigate the field more effectively, contributing to the development of reliable, safe, and robust AI systems.
\end{abstract}



\end{frontmatter}



\section{Introduction}

Supervised machine learning classification involves training a model on a set of labeled examples and then deploying it to predict these classes for new instances of input. Typically, it operates under a closed-set assumption, where the labels (classes) are known in advance and match the training distribution. This assumption simplifies learning, but without specific out-of-distribution (OOD) mechanisms, the model cannot identify inputs that are significantly different from the training data.  
OOD refers to inputs that differ significantly from the inputs used to train a model. 
These inputs may include:
\begin{itemize}
    \item unseen classes, which are not present in the training set. For example, training datasets for autonomous driving typically consist of common urban objects like cars and pedestrians, so animals not seen in urban environments (like kangaroos) may be an OOD object.
    \item anomalous samples, where the object is in-distribution, however, has previously unseen characteristics, like a vehicle with a large bullbar, trailer, or other modification which is not prevalent in the training set; or
    \item appearance changes, which may arise, for example, when all objects in the training dataset are captured in a specific style, while their appearance may change during inference due to different lighting or weather conditions.
\end{itemize}

OOD detection plays a crucial role in enabling models to identify such inputs. It is closely related to several other tasks, including anomaly detection (AD), novelty detection (ND), open-set recognition (OSR), and outlier detection (OD). \cite{yang2024generalized} collectively define these tasks within the framework of Generalized Out-of-Distribution Detection.

Distribution shift describes the changes between the training data distribution and the test/deployment distribution of a machine learning model. Distribution shift can be grouped into two categories: semantic shift and co-variate shift, which corresponds respectively to the tasks of novel input detection (OOD detection) and anomalous input detection (anomaly detection).

To explain semantic shift and covariate shift, we define the input space as $\mathbf{X}$ (sensory observations) and the label space as $\mathbf{Y}$ (semantic categories) \citep{yang2024generalized}. The data distribution is represented by the joint distribution $P(\mathbf{X}, \mathbf{Y})$ over $\mathbb{R}^{(\mathbf{X} \times \mathbf{Y})}$. 
Semantic shift involves changes in the label distribution $P(\mathbf{Y})$, often introducing new categories or altering existing ones. These label changes also alter the input distribution $P(\mathbf{X})$, as the observed data reflect the modified labels in the joint distribution.
For example, in autonomous driving, an OOD detector might be used to identify new object categories that were not part of the training set. In contrast, covariate shift only affects $P(\mathbf{X})$, manifesting as input distortions, such as corruption or stylistic variations.

Within the broader landscape of OOD research, techniques like Domain Adaptation (DA) \citep{dasurvey18neurocomp} and Domain Generalization (DG) \citep{dgsurvey21arxiv} represent specialized approaches to handling OOD scenarios. While these methods operate under an "open world" assumption, they address fundamentally different challenges compared to OOD detection. DA/DG techniques specifically target covariate 
shift in data distribution while maintaining the same semantic categories. In contrast, OOD detection predominantly focuses on identifying semantic shift, where entirely new classes or concepts emerge that were not present during training. This survey specifically examines OOD detection methods, leaving OOD generalization approaches like DA/DG outside its scope.


OOD detection plays a crucial role in enhancing the trustworthiness of machine learning models, attracting significant research interest, leading to numerous related studies.
Early research on OOD detection is rooted in novelty detection \citep{ndsurvey03sp01,ndsurvey03sp02} and open set recognition \citep{towardosr13pami}, both of which focus on identifying unknown inputs in shallow domains, where the unknowns are still valid inputs that resemble the style of the training data. First introduced by \cite{msp}, OOD detection aims to identify unknown inputs in open domains, where the test set contains highly diverse unknown data.

\begin{table*}[h]
    \centering
        \caption{An overview of existing OOD-related survey papers. Generalized Out-of-distribution Detection is a collective concept of out-of-distribution detection (OOD), anomaly detection (AD), novelty detection (ND), open-set recognition (OSR), and outlier detection (OD). OOD generalization tasks, such as domain adaptation and domain generalization, are out of the scope of this survey.}
    \begin{adjustbox}{width=\columnwidth}
    \begin{tabular}{lclll}
        \hline
        \textbf{Reference} & \textbf{Year} & \textbf{Research Topic} & \textbf{Data Modality} & \textbf{3D Application} \\ 
        \hline
        \cite{boult19aaai} & 2019 & Open World Recognition & RGB Image & Not Covered \\ 
        \cite{anomalysurvey19sydney} & 2019 & Anomaly Detection & RGB Image & Not Covered \\ 
        \cite{baul18} & 2019 & Anomaly Detection (Medical) & MR Image & Medical Diagnosis \\
        \cite{osrsurvey20pami} & 2020 & Open Set Recognition & RGB Image & Not Covered  \\
        \cite{anomalyreview20adelaide} & 2021 & Anomaly Detection & RGB Image, Video, etc. & Not Covered \\ 
        \cite{unifiedood} & 2022 & Generalized Out-of-distribution Detection & RGB Image & Not Covered \\ 
        \cite{Bogdoll_2022} & 2022 & Anomaly Detection (Driving Scene) & RGB Image & Not Covered \\
        \cite{Bogdoll_2023} & 2023 & Anomaly Detection (Driving Scene) & RGB Image & Not Covered \\
        \cite{yang2024generalized} & 2024 & Generalized Out-of-distribution Detection & RGB Image & Not Covered \\ 
        \cite{liu_industrial_survey} & 2024  & Anomaly Detection (Industrial)  & RGB Image& Not Covered \\
        \cite{Rani_2024}  & 2024  & Anomaly Detection (Industrial) & Point Cloud & Defect Detection \\
        \cite{Liu_2024_industrial_ad} & 2024  & Anomaly Detection (Industrial) & RGB Image, Point Cloud & Defect Detection \\
        Ours & 2025 & Generalized Out-of-distribution Detection & Various forms of 3D Data  & Various 3D Applications \\
        \hline
    \end{tabular}
    \label{tab:survey_comparison}
    \end{adjustbox}
    \end{table*}

\begin{figure*}
  \includegraphics[width=\textwidth]{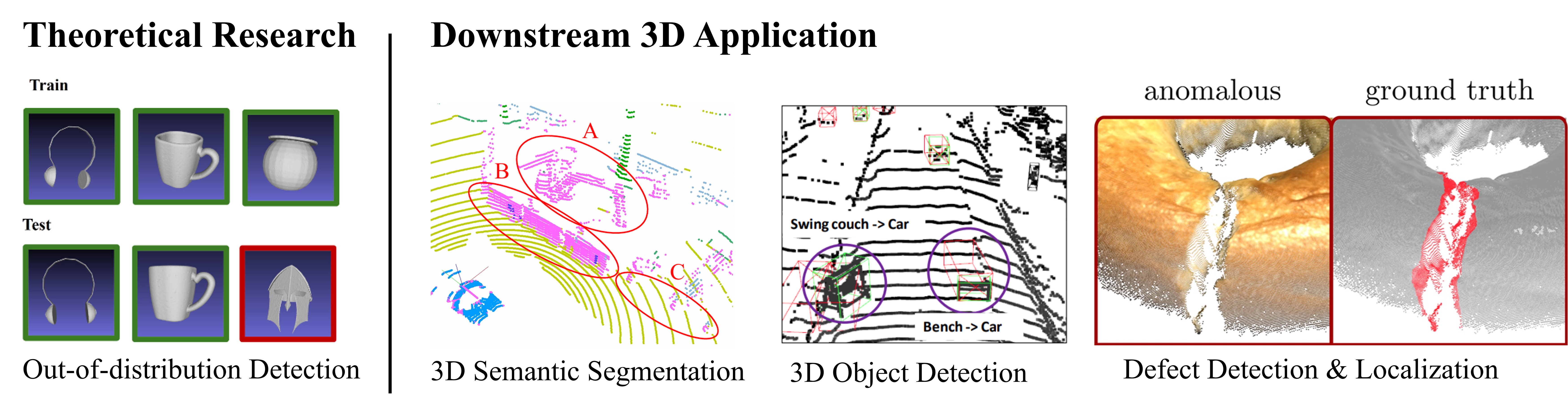}
  \caption{Generalized OOD detection and its potential 3D applications.}
  \label{fig:ood43d}
\end{figure*}

While several survey papers on OOD detection exist, as shown in \cref{tab:survey_comparison}, these works generally focus on the theoretical aspects of OOD detection \citep{yang2024generalized,boult19aaai,osrsurvey20pami,osrsurvey21arxiv,ndsurveyox14sp,ndreview10mipro,ndsurvey03sp01,ndsurvey03sp02,unifiedood} and anomaly detection \citep{anomalysurvey21ieee,anomalyreview20adelaide,anomalysurvey20dsong,anomalysurvey19sydney} predominantly in 2D image classification, or concentrate on a specific downstream application \citep{baul18,Bogdoll_2022,lin2024surveyrgb3dmultimodal,liu_industrial_survey,Bogdoll_2022,Rani_2024}. A comprehensive survey dedicated to 3D OOD detection covering various downstream applications, sensor modalities, and providing insightful methodological discussions—is still lacking. 
As shown in \cref{fig:ood43d}, 3D OOD detection refers to the task of identifying unknown or anomalous objects in three-dimensional data representations, such as point clouds, voxel grids, and depth maps. Unlike traditional OOD detection, which primarily focuses on 2D images, 3D OOD detection must account for spatial geometry, structural variations, and sensor noise. This is particularly important in applications like autonomous driving, robotics, and medical imaging.

This paper aims to bridge the gap by reviewing recent advancements in 3D OOD detection. We provide a comprehensive overview of key methodologies, benchmark datasets, and practical applications across domains such as autonomous driving, industrial defect detection, and medical diagnosis. We also discuss open challenges in 3D OOD detection and identify opportunities for future research.

\section{Timeline of the Development of OOD Detection}

\begin{figure*}
  \includegraphics[width=\textwidth]{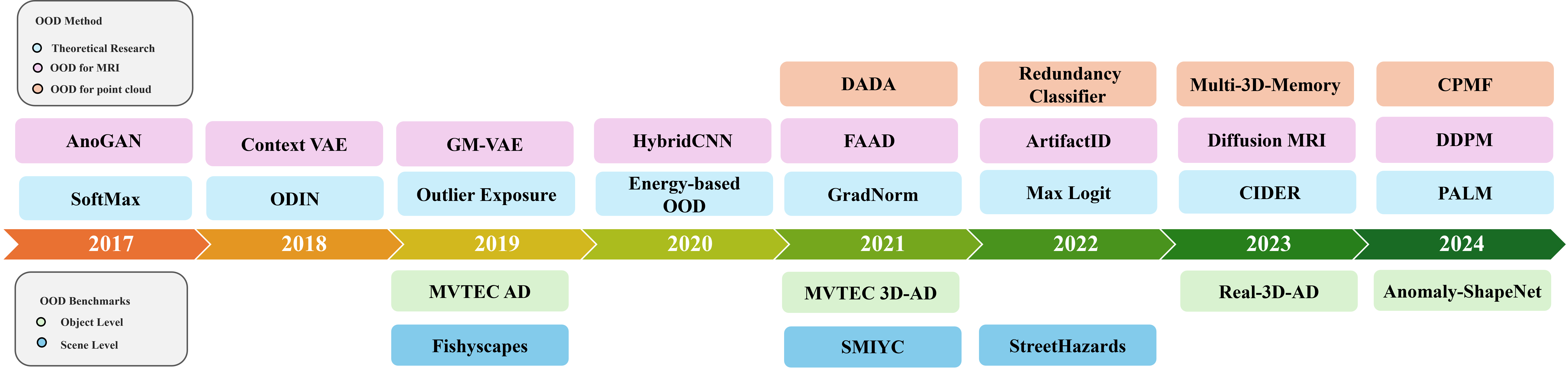}
  \caption{Timeline of the development of OOD detection. We include representative OOD detection methods and benchmark datasets based on their influence. Algorithms are grouped with different colours based on their methodologies and data modalities.}
  \label{fig:timeline}
\end{figure*}

As shown in \cref{fig:timeline}, since 2017, OOD detection has emerged as a critical research area in machine learning, focusing on identifying unknown inputs in open domains where the test set contains highly diverse unknown data.
The research landscape of OOD methods has witnessed significant transformations from 2017 to 2024, characterized by innovative techniques across theoretical research, medical imaging, and point cloud application domains.

\noindent\textbf{Methodological Progression.} The methodological evolution began with simple approaches like SoftMax \citep{advopenmax17bmvc} and Mahalanobis Distance \citep{denouden2018improving} in 2017-2018, progressively advancing to more complex techniques. Generative models such as Anomaly detection GAN \citep{adgan18ecml} and Context VAE \citep{denouden2018improving} demonstrated early potential in anomaly detection. By 2019-2020, methods like Outlier Exposure \citep{oe}, Energy-based OOD detection \citep{energy}, and Contrastive Shift Instance \citep{csi} emerged, which significantly improved OOD detection performance by involving auxiliary data. The breakthrough period from 2021-2023 saw the introduction of advanced techniques like Depth-Aware Discrete Autoencoders \citep{
zhang2020uc}, Multi-3D Memory \citep{pmlr-v202-chu23b} and Adversarial Prototypes \citep{apf23}, and complex probabilistic model \citep{bohm2020probabilistic}, for downstream applications. At the same time, theoretical research is also deepening, emerging influential works such as  GradNorm \citep{huang2021importance}, Max Logit \citep{maxlogit}, CIDER \citep{cider}, and LogitNorm \citep{wei2022mitigating}, which provide increasingly refined mechanisms for distinguishing between in-distribution and out-of-distribution data samples.

\noindent\textbf{Data-Modality-Specific Advancements.} Specialized research threads emerged across different domains. For medical imaging (MRI), techniques like Diffusion MRI \citep{weninger2023diffusion} and FAAD \citep{faad21} demonstrated significant progress in anomaly detection for medical diagnosis. Point cloud research witnessed innovations such as PointAD \citep{zhou2024pointad}, CPMF \citep{cpmf} and PDF \citep{pdf24}, addressing the unique challenges of 3D and multimodal anomaly detection.

\noindent\textbf{Benchmark Developments.} Benchmark datasets evolved correspondingly with OOD methods, reflecting the focus change of the research community. Initial benchmarks like Fishyscapes \citep{fishyscapes} in 2019 were succeeded by more comprehensive datasets such as SMIYC \citep{smiyc}. Large-scale driving scene perception datasets such as SemanticKitti \citep{SemanticKITTI} and nuScenes \citep{nu} are also reused for OOD detection tasks.  The progression from MVTEC AD \citep{mvtec19cvpr} to Real3D AD \citep{liu2023real3d} and Anomaly-ShapeNet \citep{ashapenet} illustrates the increasing complexity and domain-specific OOD evaluation frameworks from 2D to 3D.

\noindent\textbf{Emerging 3D Application.}
In recent years, significant progress has been made in 3D OOD detection \citep{weninger2023diffusion, gawlikowski2022advanced}. Notably, advancements in industrial anomaly detection \citep{realiad, bhunia2024look3d, liu2023real3d, bergmann2021mvtec, li2024multisensorobjectanomalydetection} and autonomous driving \citep{oodnogpu, li2024contrastive} have underscored the crucial role of accurate OOD identification in ensuring safety. The field continues to push the boundaries of traditional machine learning paradigms, enhancing model robustness across diverse domains and increasingly complex real-world 3D scenarios.

\begin{table*}[h!]
\centering
\caption{A taxonomy of OOD detection research. We will first introduce the OOD detection from the perspective of downstream 3D applications. Then we will discuss how OOD detection has been implemented for different forms of 3D data. Finally, we summarize OOD detection methodologies that can be applied to 3D applications.}
\label{tab:four_columns}
\tiny
\setlength{\tabcolsep}{6pt} 
\renewcommand{\arraystretch}{1.2} 
\begin{adjustbox}{width=\columnwidth}
\begin{tabular}{>{\raggedright\arraybackslash}m{1.5cm}|>{\raggedright\arraybackslash}m{2cm}|>{\raggedright\arraybackslash}m{1.5cm}|>{\raggedright\arraybackslash}m{10cm}}
\toprule
\multicolumn{3}{l|}{\textbf{Topic}} & \textbf{Reference} \\ \hline

\multirow{12}{*}{Application}        & \multirow{5}{*}{Autonomous Driving} & Segmentation                 & \citep{msp,md,dense,pweal,rpl,metaood,pweal,sleeg,owss,lis,synboost,maxlogit,deli2024outlier,rba23,Grci__2023,Maskomaly23,Rai_2023,mcdropout16icml,apf23} \\ \cmidrule(l){3-4} 
        & & Object Detection                 & \citep{msp,md,revisit2024,oodnogpu,itsc22,gal,ocsvm,normalizingflows,du2022vos,li2024open,liu2024yolouniow}  \\ \cmidrule(l){2-4} 
        & \multicolumn{2}{l|}{Industrial}                   & \citep{zavrtanik2021draem,cdo,RudWeh2023,bauza2019omnipush,nguyen2022effective,shi2022learning,kong2024robodepth,wang2023multimodal,pmlr-v202-chu23b,nif}  \\ \cmidrule(l){2-4} 
        & \multicolumn{2}{l|}{Medical}             &  \citep{heer2021ood,pmlr-v102-you19a,chen2018unsupervised,heer2021ood,pmlr-v102-you19a,baul18,BAUR2021101952,context,larsen2016autoencoding,baul18}  \\ \cmidrule(l){2-4} 
        & \multicolumn{2}{l|}{Remote Sensing}                 & \citep{hyperosr,gawlikowski2022advanced,fewshothyper21,pal2021,inkawhich2022improving}  \\ \hline
\multirow{12}{*}{Sensor}    & \multirow{8}{*}{Point Cloud} &   LiDAR              &  \citep{msp,md,ocsvm,normalizingflows,itsc22,revisit2024,owss,oodnogpu,apf23,pdf24}   \\ \cmidrule(l){3-4} 
        &  &   Radar              & \citep{inkawhich2022improving,odin18iclr,kahya2023mcrood,Griebel21}   \\ \cmidrule(l){3-4} 
        &  &   Industrial Senor              & \citep{cdo,liu2023real3d,wang2023multimodal,pmlr-v202-chu23b,nif,cdo,RudWeh2023,bauza2019omnipush,nguyen2022effective,shi2022learning,kong2024robodepth,zhou2024pointad,btf,zhou2024r3dad,ashapenet}   \\ \cmidrule(l){2-4} 
& \multicolumn{2}{l|}{MR Imaging}                 & \citep{heer2021ood,pmlr-v102-you19a,chen2018unsupervised,heer2021ood,pmlr-v102-you19a,baul18,BAUR2021101952,context,larsen2016autoencoding}   \\ \cmidrule(l){2-4} 
        & \multicolumn{2}{l|}{Sensor Fusion} & \citep{cdo,RudWeh2023,bauza2019omnipush,nguyen2022effective,shi2022learning,kong2024robodepth,wang2023multimodal,pmlr-v202-chu23b,nif} \\ \hline
        \multirow{20}{*}{Method}         & \multirow{8}{*}{Logit-based Method}           & Training-free            & \citep{msp,energy,mcm,clipn,maxlogit,openmax,jenergy,mood21cvpr,djurisic2022extremely,park2023nearest,jiang2023detecting,liu2023gen} \\ \cmidrule(l){3-4} 

&             &    Training-based         & \citep{energy,wei2022mitigating,oe,odin18iclr,energy,devries2018learning,wang2021energy,eloc18eccv,good20nips,aloe20arxiv,ceda19cvpr,blur20iclr,outliermining21ecml,mixup19nips,cutmix19cvpr,cutout17arxiv,augmix19arxiv,hendrycks2021pixmix,ccu20arxiv,bibas2021single,wang2022watermarking,mood21cvpr,godin20cvpr,hierarchical18cvpr,mos21cvpr,hierarchyood,nearood21arxiv,cac,agnostophobia18nips} \\ \cmidrule(l){2-4} 
            
           & \multirow{5}{*}{Feature-based Method}           & Training-free           &  \citep{md,maxlogit,gram20icml,sun2021tone,dongneural,sun2022dice,knn22,gram19nipsw,park2023understanding,vim,she23,deepsvdd18icml} \\ \cmidrule(l){3-4} 
        &            & Training-based           & \citep{gram20icml,knn22,csi,dongneural,sehwag2021ssd,ksemantic18nips,cider,PALM2024,li2024contrastive,deepsvdd18icml,tao2023nonparametric,xu2024scaling_ish,mirzaei2024adversarially} \\ \cmidrule(l){2-4} 
        
        & \multicolumn{2}{l|}{Reconstruction-based Method}                  & \citep{dense,vae13arxiv,vqvae,Schlegl_2017,SCHLEGL201930,Gong_2019_ICCV,c2ae19cvpr,chen2018unsupervised,heer2021ood,pmlr-v102-you19a,baul18,BAUR2021101952,zavrtanik2021draem} \\  \cmidrule(l){2-4}
       & \multicolumn{2}{l|}{Generative Method} & \citep{confcal18iclr,oodsg19nipsw,gopenmax,confgan18nipsw,maml20nips,auroc,du2022vos,du2022unknown,Schlegl_2017,SCHLEGL201930,npos2023iclr,wang2023out,zheng2023out} \\
\bottomrule
\end{tabular}
\end{adjustbox}
\end{table*}



Although OOD detection has received significant attention in the context of 2D data, particularly in image classification, extending these methods to 3D data presents unique complexities.
One key challenge in 3D OOD detection is the inherent sparsity and incompleteness of real-world 3D scans. Unlike 2D image data, which is typically represented as pixel arrays with fixed dimensions, 3D data can take various forms, including point clouds, voxel grids, and meshes. These representations introduce additional complexities such as irregular sampling, varying density, and spatial transformations \citep{3dptsurvey}. This variability makes it more difficult to establish a clear boundary between in-distribution and out-of-distribution samples. Additionally, 3D objects exhibit significant intra-class variation due to viewpoint changes, scale differences, and deformations, further complicating the task.


Despite these complexities, 3D OOD detection also presents unique opportunities. The geometric nature of 3D data provides additional information that can enhance the robustness of OOD detection models. Techniques such as shape-based feature learning \citep{ashapenet}, contrastive learning \citep{li2024contrastive}, and knowledge distillation \citep{RudWeh2023} can help improve OOD detection performance in 3D domains. Additionally, multi-modal approaches that combine 2D and 3D data can further enhance system reliability, especially in safety-critical applications like autonomous driving and medical diagnostics.

In \cref{tab:four_columns}, we present a taxonomy of OOD detection research that outlines the scope and structure of this paper.
In terms of applications, OOD detection has been extensively studied in autonomous driving (for both segmentation and object detection), as well as in industrial, medical, and remote sensing settings. The research spans various sensor types, including LiDAR, radar, MR imaging, and industrial sensors, with increasing attention to sensor fusion techniques. Methodologically, the field is broadly divided into logit-based, feature-based, reconstruction-based, and generative approaches, each further split into training-free and training-based paradigms.
We will first introduce the OOD detection from the perspective of downstream 3D applications in \cref{sec:app}. Then we will discuss how OOD detection has been implemented for different forms of 3D data in \cref{sec:sensor}. Finally, we introduce OOD detection methodologies that can be applied to 3D applications in \cref{sec:method}.


\begin{table}[!thbp]
\centering
\caption{\small{Summary of evaluation datasets.}}
\begin{adjustbox}{width=\columnwidth}
\begin{tabular}{c c c c}
\toprule
  Dataset  &  Data source & Data type  &  Application \\ 
  \hline
  Lost and Found \citep{lostandfound} & Real          & RGB Image       & Autonomous Driving \\ 
  Fishyscapes \citep{fishyscapes}   & Real + Synthetic          & RGB Image      & Autonomous Driving\\
  SMIFC \citep{smiyc}   & Real          & RGB Image &   Autonomous Driving \\ 
  StreetHazards \citep{maxlogit}      & Synthetic       & RGB Image       & Autonomous Driving \\ 
  nu-OWODB \citep{li2024open}   & Real          & RGB Image &   Autonomous Driving \\ 
  KITTI \citep{kitti} & Real & RGB + Point Cloud & Autonomous Driving \\
  SemanticKITTI  \citep{SemanticKITTI} & Real & Point Cloud & Autonomous Driving \\
  nuScenes \citep{nu} & Real & RGB + Point Cloud & Autonomous Driving \\
  \hline
  MSSEG2015 \citep{carass2017longitudinal} & Real          & MR Image       & Lesion Detection \\
  MSLUB \citep{MSLUB} & Real          & MR Image       & Lesion Detection \\
  MR-ART \citep{narai2022movementmr-art} & Real          & MR Image       & Lesion Detection \\
  \hline
  SAMPLE \citep{SAMPLEradar} & Real + Synthetic & Radar Data & Remote Sensing \\
  MLRSNet \citep{MLRSNet} & Real & RGB Image & Remote Sensing \\
  \hline
  MVTec AD \citep{mvtec19cvpr} & Real & RGB Image & Industry \\
  3CAD \citep{yang3cad} & Real & RGB Image & Industry \\
  MVTec 3D-AD \citep{bergmann2021mvtec} & Real & Point Cloud + RGB Image & Industry \\
  Real 3D-AD \citep{liu2023real3d} & Real & Point Cloud & Industry \\
  Anomaly-ShapeNet \citep{ashapenet} & Synthetic & Point Cloud & Industry \\ 
   Real-IAD \citep{realiad} & Real & Multi-View RGB & Industry \\
   Looking 3D \citep{bhunia2024look3d} & Synthetic  & Multi-View RGB & Industry \\
   MulSen-AD \citep{li2024multisensorobjectanomalydetection} & Real & Image (RGB+Infrared) + Point Cloud & Industry \\
  \bottomrule 

\end{tabular}
\end{adjustbox}
\label{tab:dataset_summary}
\vspace{-0.8em}
\end{table}

\section{OOD Detection Benchmarks and 3D Applications}
\label{sec:app}
OOD detection is crucial in various real-world applications where safety and reliability are paramount. In autonomous driving, it helps vehicles recognise unknown objects or anomalous conditions, preventing accidents caused by unknown obstacles. For industrial applications, OOD methods identify anomalous manufacturing defects that deviate from standard product specifications, ensuring quality control. In medical imaging, detecting OOD samples aids in flagging rare diseases or unseen abnormalities that may require further clinical assessment. Similarly, in remote sensing, OOD detection enhances land cover classification and disaster monitoring by identifying unexpected environmental changes or novel structures in satellite imagery. As shown in \cref{tab:dataset_summary}, we provide a summary of datasets commonly used in prior studies on real-world OOD detection, highlighting their data sources, data types, and application domains. 
These datasets vary in data source (real, synthetic, or a combination), modality (e.g., RGB images, point clouds, MR images, radar, infrared), and task specificity. Notably, autonomous driving datasets dominate the landscape, often leveraging real-world RGB or LiDAR data, while industrial and medical domains increasingly incorporate multi-modal or 3D data for more robust anomaly detection. This diversity underscores the growing demand for versatile benchmarks to evaluate models' OOD performance.
In the following subsections, we describe how OOD detection is performed across different downstream applications.

\subsection{Autonomous Driving}
In autonomous driving, OOD detection is essential for ensuring the safety of vehicles in dynamic and unpredictable environments. Autonomous systems rely on machine learning models trained on specific datasets that represent common road conditions, objects, and behaviours. However, real-world driving scenarios often introduce unexpected road objects, such as debris, animals, unfamiliar vehicle types and road incidents \citep{incidents}, that fall outside the training distribution. OOD detection enables autonomous systems to recognize and respond appropriately to these unexpected inputs, ensuring safe navigation and decision-making.

This presents a scene understanding challenge that not only involves detecting OOD objects or anomalies but also requires their segmentation. For image-based inputs, there has been a significant body of research on OOD detection and anomaly segmentation in autonomous driving, along with several benchmarks \citep{maxlogit,lostandfound,fishyscapes,smiyc} for outdoor semantic segmentation.

Lost and Found \citep{lostandfound} focuses on detecting small obstacles on the road. Fishyscapes \citep{fishyscapes} introduces out-of-distribution (OOD) objects into driving datasets such as Cityscapes \citep{cityscapes}. It also filters scenes in Lost and Found \citep{lostandfound} to retain in-distribution objects, making it suitable for models trained on Cityscapes. SegmentMeIfYouCan (SMIYC) \citep{smiyc} is similar to Fishyscapes \citep{fishyscapes}, but it extends the diversity of objects and evaluation metrics. StreetHazards \citep{maxlogit} is a synthetic dataset based on the CARLA simulator \citep{carla}. Using a simulated environment allows for the dynamic placement of a wide range of anomalous objects in various locations while ensuring visual coherence with real-world settings.


Some OOD detection methods \citep{oe,dense,pweal,rpl,mask2anomaly} utilize an Outlier Exposure (OE) \citep{oe} strategy, which uses objects from other datasets as auxiliary outlier samples. The vanilla OE \citep{oe} optimizes the model by forcing it to output uniform logits for outlier samples. \cite{metaood} propose Meta-OOD, which crops objects from the COCO dataset \citep{coco} to the CityScapes dataset \citep{cityscapes} and maximizes the model's softmax entropy to outlier pixels. Similarly, \cite{pweal} propose Pixel-Wise Energy-Biased Abstention Learning, and Energy Based Model \citep{ebm}, which produces high energy for outlier pixels.

However, retraining the model with the OE dataset will also shift the decision boundary of known classes \citep{sleeg,rpl}. \cite{rpl} propose Residual Pattern Learning (RPL), which is a trainable module that connects the encoder and the decoder of the semantic segmentation model. The RPL block is the only trainable module specialized for out-of-distribution detection; all other modules are frozen so that decision boundaries for known classes are fixed. In addition, \cite{rpl} propose Context-robust Contrastive Learning (CoroCL), which pulls embeddings in the same class together while pushing away embeddings from other classes and outlier classes. Treating anomaly detection as a per-pixel classification task will lead to high uncertainty at object boundaries and numerous false positives. \cite{mask2anomaly} propose Mask2Anomaly, a mask classification framework that incorporates several innovations: a global masked attention module to focus separately on the foreground and background, mask contrastive learning to maximize the separation between anomalies and known classes, and a mask refinement solution to reduce false positives.

Image resynthesis typically uses GAN \citep{gan} or autoencoder \citep{gen} to create new images based on certain conditions. \cite{lis} observe that segmentation models produce spurious labels in areas showing out-of-distribution objects. As a result, resynthesizing the image from such a semantic map will lead to significant visual differences compared to the original image. In this way, image resynthesis transforms the task of detecting unknown classes into identifying regions where the image has been poorly resynthesized. \cite{synboost} propose SynBoost, which uses an additional spatial-aware dissimilarity network to simultaneously detect segmentation uncertainty and image resynthesis error. These methods are self-supervised, without the need for auxiliary out-of-distribution data. However, these approaches largely depend on the quality of segmentation maps and performance of the image resynthesis model, In addition, their effectiveness can be compromised by artifacts generated by GANs \citep{sleeg}.

LiDAR is promising for OOD detection in autonomous driving due to its high spatial resolution, which provides detailed 3D point cloud data, and its robustness to varying lighting conditions. However, the research related to LiDAR-based OOD is relatively limited, and there is no dedicated benchmark dataset. The evaluation is usually conducted on public autonomous driving datasets such as KITTI \citep{kitti}, nuScenes \citep{nu}, and the CARLA simulator \citep{carla}.

For 3D object detection, \cite{itsc22} evaluate various post-hoc OOD detection methods on top of the Pointpillars \citep{pointpillars} backbone, synthesizing OOD samples from real and simulated datasets due to a lack of dedicated datasets. \cite{revisit2024} highlight potential issues with synthetic evaluations and propose using the nuScenes dataset for more realistic assessments. \cite{oodnogpu} introduce a real-time energy-based OOD detection framework using PointNets, achieving high processing speeds. Other approaches, such as REAL \citep{owss} and APF \citep{apf23}, focus on open set semantic segmentation by integrating redundancy classifiers and adversarial techniques.

\subsection{Industrial Applications}
In industrial applications, out-of-distribution (OOD) detection ensures safety, efficiency, and quality control. It helps identify anomalies such as novel defects in manufacturing processes, unexpected deviations in machinery performance, or abnormal conditions in automated systems. 
It has attracted considerable attention, with various benchmarks (as shown in \cref{tab:dataset_summary}) \citep{mvtec19cvpr,realiad,bhunia2024look3d,liu2023real3d,ashapenet,bergmann2021mvtec,li2024multisensorobjectanomalydetection}.

\begin{figure}[t]
    \centering
    \includegraphics[width=0.95\linewidth]{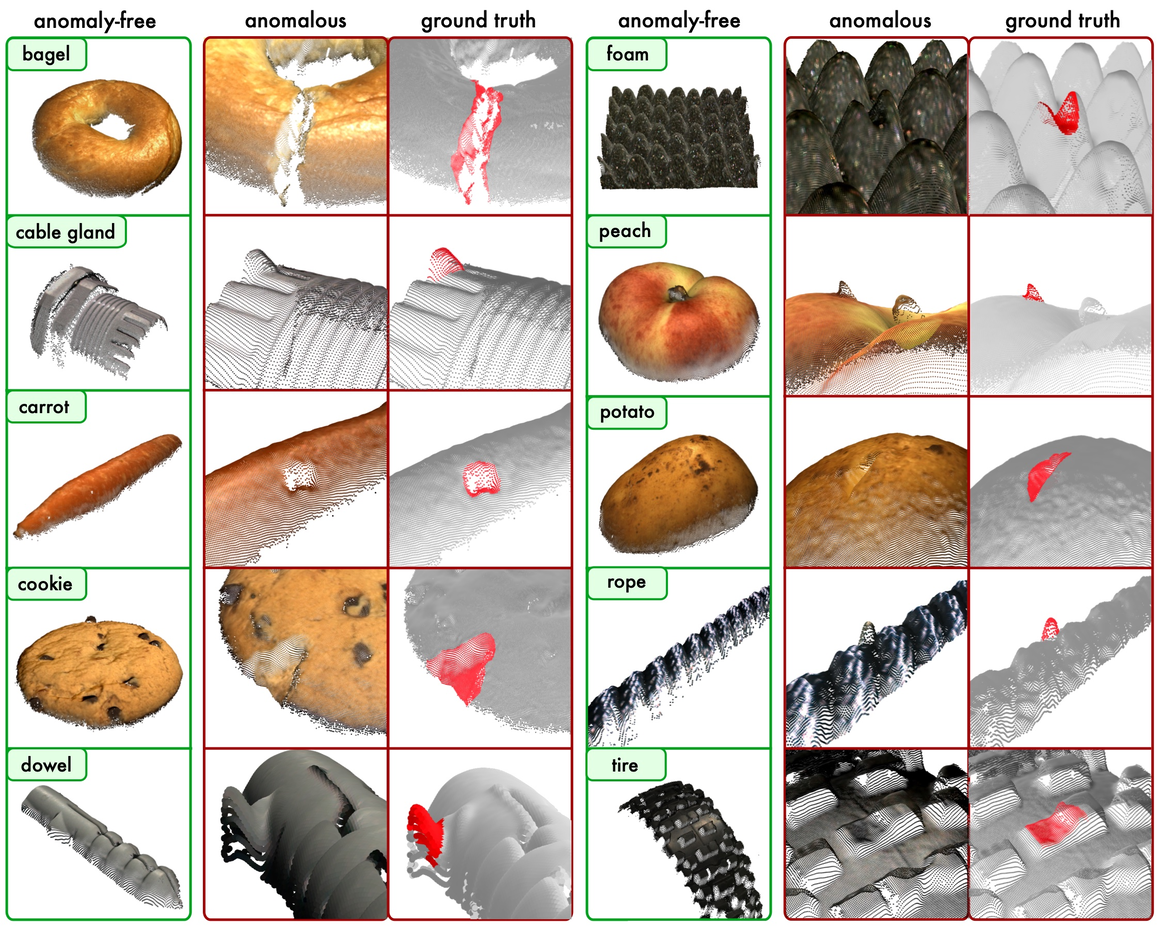}
    \caption{Examples of normal and anomalous objects from the MVTec 3D-AD dataset \citep{bergmann2021mvtec}, with anomalous regions highlighted in red.}
    \label{fig:mvt3d}
\end{figure}

In particular, 3D data modalities like point clouds \citep{liu2023real3d} and meshes \citep{ashapenet} have become increasingly prominent in OOD detection due to their ability to capture rich geometric and spatial information. Furthermore, the integration of 3D data with other modalities, such as RGB imagery \citep{bergmann2021mvtec} or infrared imaging \citep{li2024multisensorobjectanomalydetection}, enhances the robustness of OOD systems, enabling the detection of subtle anomalies that might be overlooked in single-modality analysis. 
As shown in \cref{fig:mvt3d}, a multi-modal dataset such as MVTEC 3D-AD \citep{bergmann2021mvtec} provides 3D scans with precise geometric information, for detecting anomalous surface defects and structural irregularities in objects. The dataset provides annotations not only at the object level but also at the point level, which facilitates the localization of anomalies.


\subsection{3D Medical Imaging}
In healthcare, 3D medical imaging modalities such as CT (Computed Tomography), MRI (Magnetic Resonance Imaging), and ultrasound provide detailed volumetric representations of internal organs. OOD detection for 3D medical data is critical for identifying rare or previously unseen abnormalities that may not be present in the training dataset.
For example, unsupervised lesion detection can be framed as an anomaly segmentation problem \citep{heer2021ood,pmlr-v102-you19a}, where the reconstruction error of the autoencoder is frequently involved \citep{chen2018unsupervised,heer2021ood,pmlr-v102-you19a,baul18,BAUR2021101952}. 

\begin{figure}[thbp]
  \centering
  \includegraphics[width=\linewidth]{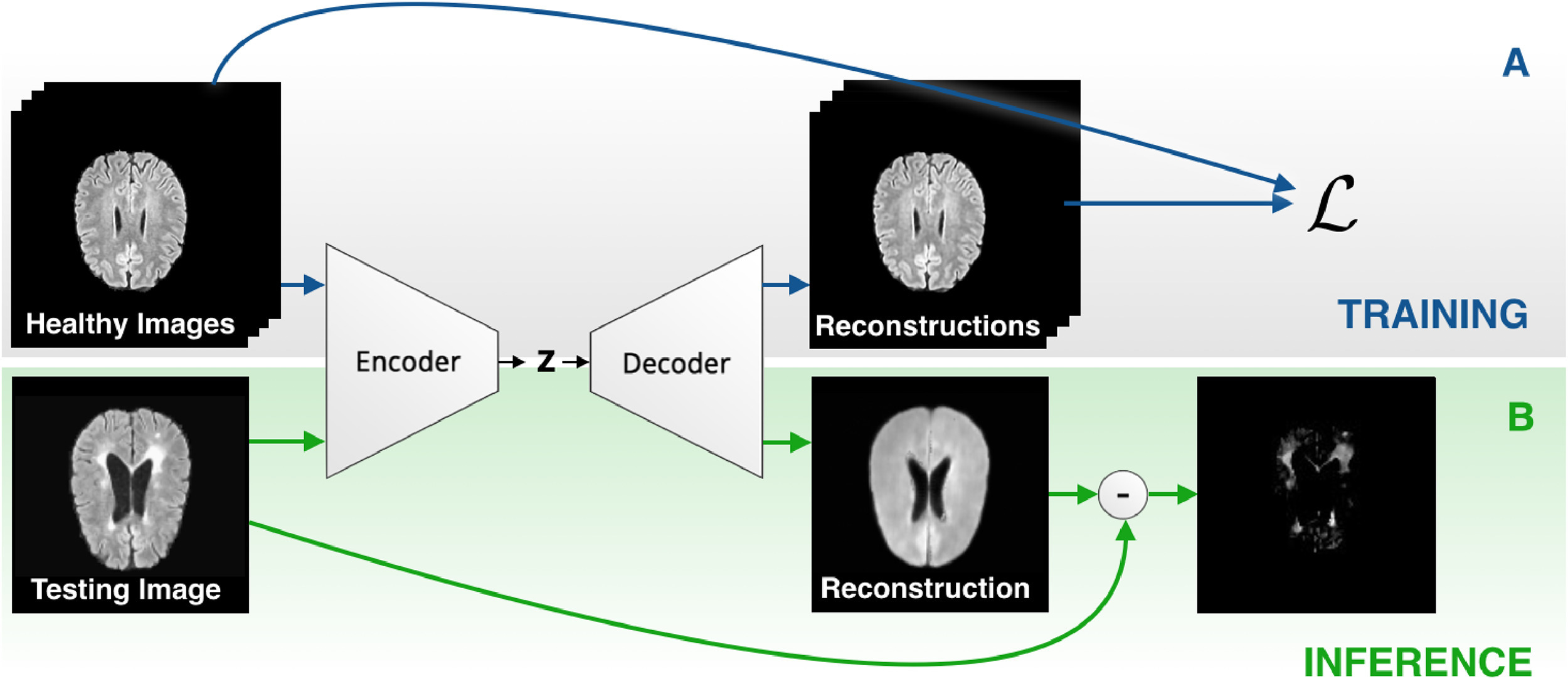}
  \caption{
Autoencoder-based OOD Detection involves: A) training a model exclusively on healthy samples, and B) identifying anomalies by segmenting regions in input samples where reconstruction errors occur, indicating potential anomalies \citep{BAUR2021101952}. }
  \label{aes2}
\end{figure}

\Cref{aes2} shows the general process anomaly segmentation with an auto-encoder. The basic idea is that the model has only been trained on healthy anatomical images, so the most probable latent representations will correspond to healthy anatomy. With this assumption, the image is reconstructed from its latent representation, resulting in large errors in the reconstruction of the lesion, while the rest of the image is accurately reconstructed. 

Anomaly scoring based on reconstruction error has two major drawbacks: it disregards the internal representation used by the model for reconstruction, and it lacks formal measures for comparing samples. To overcome these limitations, \cite{context} propose Context AE, which combines reconstruction-based and density-based anomaly scoring to improve the accuracy and consistency of anomaly detection. It is also possible to append a discriminator after a VAE to create a VAE-GAN for improved performance \citep{larsen2016autoencoding,baul18}.
In addition, to improve the structure of the autoencoder, using Monte Carlo Sampling in reconstruction provides more reliable detection of abnormalities in CT scans \citep{Pawlowski2018UnsupervisedLD}.

In addition to using autoencoders for input reconstruction, other works also use denosing diffusion \citep{weninger2023diffusion,ddpm2024}, discriminative models \citep{MANSOJIMENO202242,hybridcnn2020}, and transfer learning \citep{faad21}

\subsection{Remote Sensing}
In remote sensing applications, such as satellite image analysis for disaster monitoring or agricultural assessment, OOD detection plays a critical role in identifying anomalies, including new infrastructure, natural disasters, or other unexpected changes. However, this area has received relatively little attention. 
\cite{gawlikowski2022advanced} introduce a Dirichlet prior network-based model to quantify distributional uncertainty in deep learning-based remote sensing. Their approach enhances the representation gap between in-domain and OOD samples for improved OOD segregation at test time. 
\cite{inkawhich2022improving} adapt ODIN \citep{odin18iclr} for synthetic aperture radar imagery, evaluating it on mixed real and OOD data.
\cite{hyperosr} employ Extreme Value Machine (EVM) for open set learning in remote sensing. \cite{fewshothyper21} address the neglect of OOD samples in classification by proposing a multitask deep learning method for few-shot open set recognition. To avoid threshold-based rejection, \cite{pal2021} introduce an Outlier Calibration Network (OCN) with a residual 3D convolutional attention module, enabling threshold-free outlier prediction and data augmentation. 
While some research has been conducted \citep{hyperosr, gawlikowski2022advanced, fewshothyper21, pal2021,inkawhich2022improving}, further efforts are needed, particularly in addressing data modalities like representation learning for radar point cloud, which contains richer spatial information compared to RGB images.

\section{OOD Detection with Various Sensor Modalities}
\label{sec:sensor}
3D data can take various forms, including LiDAR point clouds, radar point clouds, and MRI scans, each possessing distinct characteristics that require specialized neural network architectures for effective processing. Given these differences, OOD detection strategies must be tailored to each modality to ensure robust performance. Moreover, multi-modal sensor fusion has emerged as a powerful approach to enhance OOD detection by integrating complementary information from different sensor modalities, including various forms of 3D data and RGB data. In this section, we explore OOD detection techniques across various sensor modalities and discuss strategies for effective multi-modal sensor fusion.
\subsection{LiDAR}
\label{sec:lidarood}
Out-of-distribution (OOD) detection is crucial for LiDAR data in several contexts, especially in safety-critical applications such as autonomous driving and robotics.

\begin{figure}[ht]
  \centering
  \includegraphics[width=\linewidth]{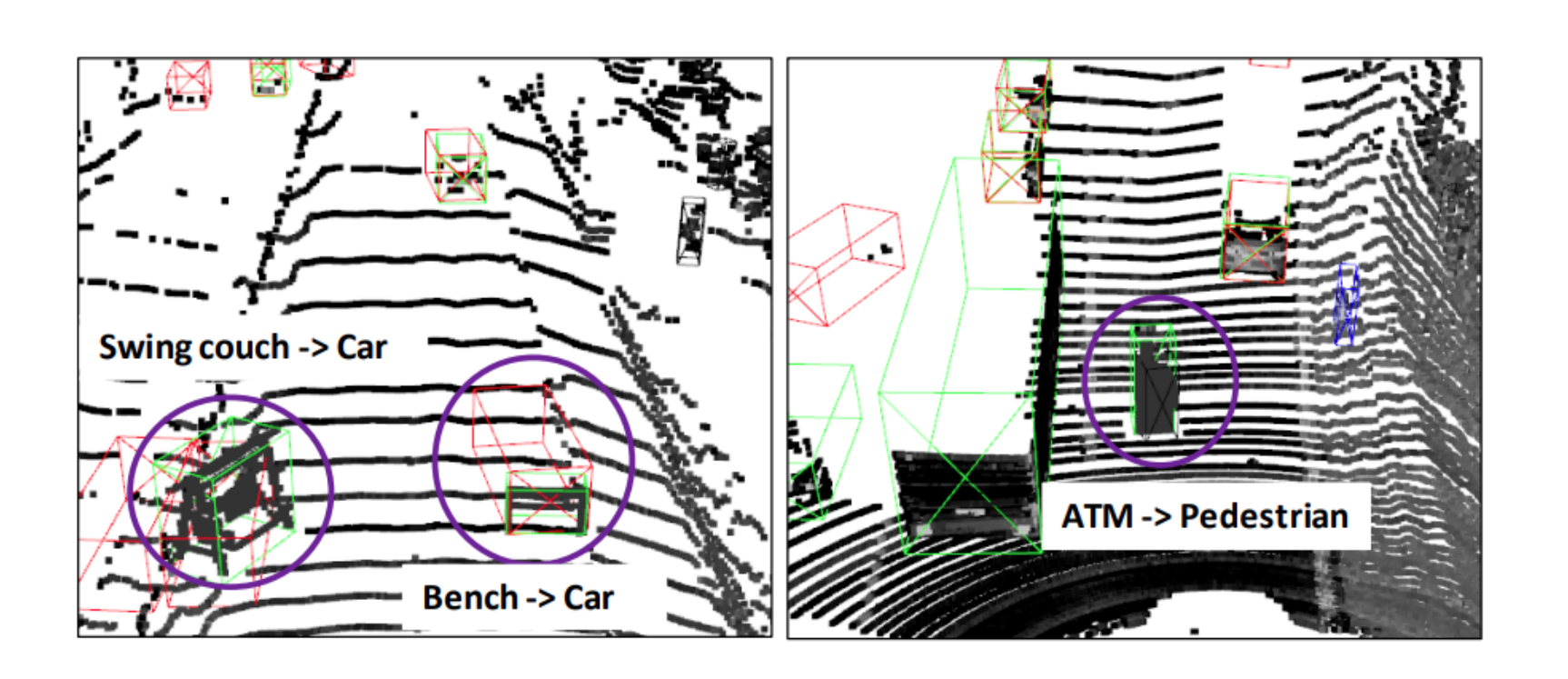}
  \caption{Injecting CARLA objects (circled in purple) into KITTI as OOD samples. Ground truth boxes are shown in green, car predictions in red, and pedestrian predictions in black. The point cloud density should be consistent to avoid the shortcut solution \citep{itsc22}.
  }
  \label{aes1}
\end{figure}

As a pioneer in this field, \cite{itsc22} evaluate five OOD detection methods: Maximum Softmax Probability (MSP) \citep{msp}, uncertainty estimation \citep{gal}, Mahalanobis Distance (MD) \citep{md}, One-Class SVM (OC-SVM) \citep{ocsvm}, and Normalizing Flows \citep{normalizingflows} in the context of a LiDAR-based 3D object detection framework. Their study employs PointPillars \citep{pointpillars} to extract 3D bounding boxes for foreground objects, followed by the application of OOD detection methods to identify OOD samples. Due to the lack of dedicated datasets for LiDAR-specific OOD detection, they synthesise OOD samples by injecting LiDAR points from real and simulated datasets into the KITTI \citep{kitti} point clouds (shown in \cref{aes1}).

While \cite{itsc22} utilize synthetic datasets for evaluation, \cite{revisit2024} argue that such synthetic approaches might introduce artificial domain gaps, such as discrepancies in intensity or a lack of natural occlusion patterns, potentially leading to unrealistic performance assessments. To overcome these limitations, they proposed using the nuScenes dataset \citep{nu}, a large-scale real-world LiDAR dataset, to improve the fidelity of OOD detection evaluations. By leveraging a pretrained object detector to generate bounding boxes, they trained an OOD detection module using synthetic OOD samples created by resizing in-distribution objects. This approach emphasizes the need for evaluations that align more closely with real-world conditions. Similarly, \cite{li2024contrastive} evaluate a variety of OOD detection methods in the nuScenes dataset and propose a novel contrastive learning based OOD method, but their work relies on ground truth bounding boxes.

Real-time performance is crucial for LiDAR-based scene understanding, because roads and environments change rapidly, requiring systems to process data and make decisions on the fly. But the inference speed is often ignored in OOD detection.
\cite{oodnogpu} introduce an energy-based OOD detection framework using PointNets for real-time 3D road user detection. Their pipeline begins with ground segmentation and clustering to generate initial proposals, which are classified by the first PointNet based on class probabilities and energy scores to filter OOD samples. The second PointNet predicts 3D bounding boxes for the remaining proposals, while a final in-distribution filter removes residual OOD instances using bounding box energy scores. A Proposal Voxel Location Encoder (PVLE) is used to preserve spatial context and retain location information during normalization. Their method achieves 76 FPS with an Nvidia GTX 1060 GPU in KITTI.
   
\begin{figure}[ht]
  \centering\includegraphics[width=\linewidth]{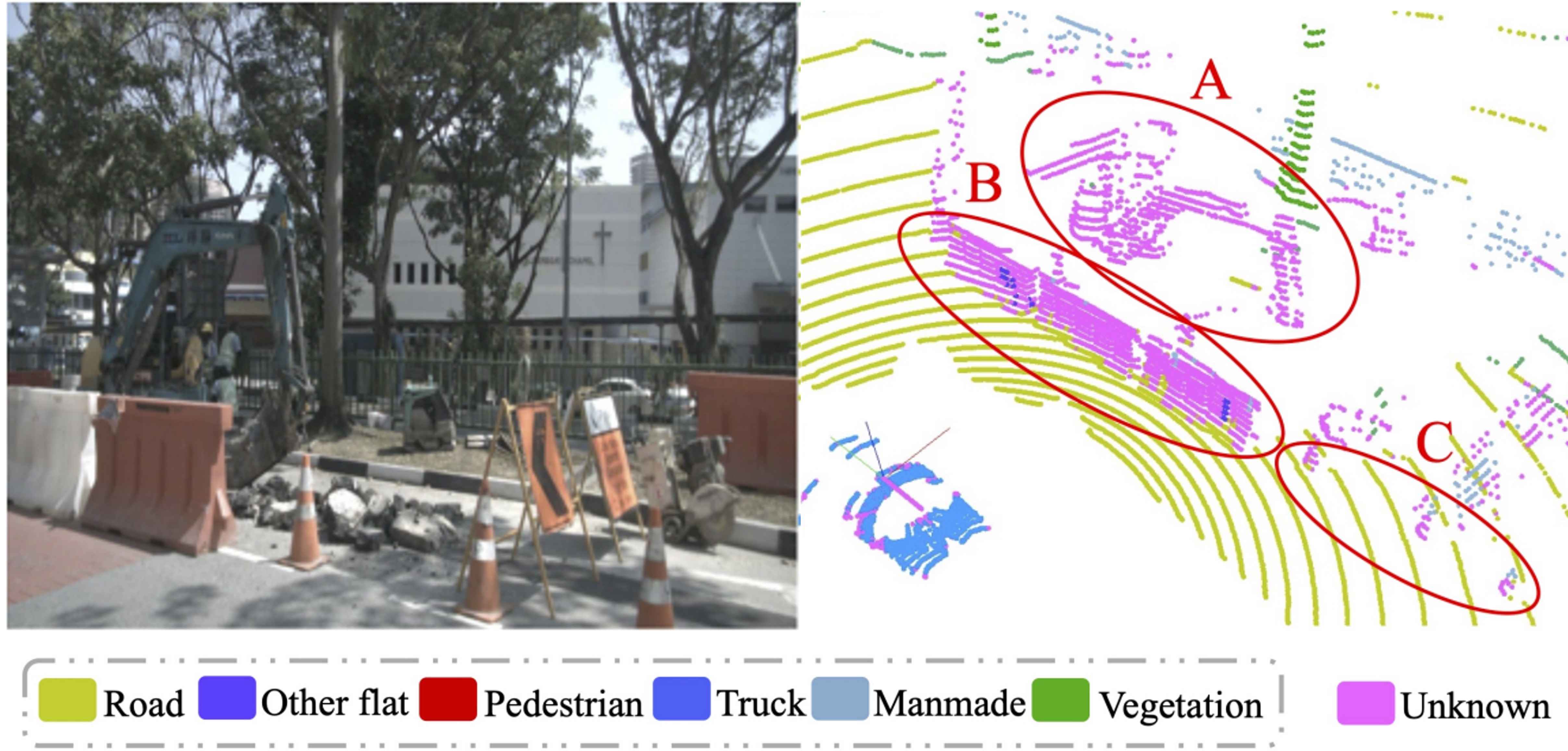}
  \caption{Visualization of open-set semantic segmentation on point clouds \citep{owss}, with point-level annotations. Purple points indicate unknown elements. A, B, and C represent the construction vehicle, barrier, and traffic cone, respectively.}
  \label{owss_vis}
\end{figure}

In 3D semantic segmentation, OOD detection involves both identifying OOD points and classifying in-distribution points in the 3D point cloud, as shown in \cref{owss_vis}.
\cite{owss} propose the REdundAncy cLassifier (REAL) framework, a dynamic architecture tailored for both open set semantic segmentation (OSeg) and incremental learning (IL) tasks. In the OSeg task, the framework enhances the original network with additional redundancy classifiers (RCs) to predict probabilities for OOD classes. REAL generates synthetic OOD point clouds by scaling existing ones and optimizes the RCs to assign higher probabilities to OOD samples. The RCs gradually learn new classes as unknown objects are annotated. Alternatively, \cite{apf23} propose the Adversarial Prototype Framework (APF) for open set semantic segmentation. The proposed APF framework comprises three key components: a feature extraction module to extract point features, a prototypical constraint module to learn prototypes for each seen class, and a feature adversarial module. The prototypical constraint module generates class-specific prototypes from the extracted features, while the feature adversarial module employs generative adversarial networks to estimate the distribution of unseen-class features. In more recent work, \cite{pdf24} introduce a lightweight U-decoder parallel to the closed-set decoder to estimate OOD-ness directly within the point cloud. 

Overall, there is still considerable potential for development in LIDAR-based 3D OOD detection, primarily due to the scarcity of diverse real-world datasets for comprehensive evaluation. Currently, OOD data styles rely heavily on synthesis and dataset repartitioning. In addition, there is a lack of significant theoretical advances in outdoor OOD detection algorithms. Most existing works focus on applying existing OOD methods to 3D point clouds, but they overlook the inherent properties of outdoor 3D point clouds, such as sparsity and occlusion.


\subsection{Radar}
For remote sensing applications, \cite{inkawhich2022improving} explore a confidence-based method ODIN \citep{odin18iclr} trained with synthetic aperture radar imagery, and evaluated with a mixture of real-world data and OOD data. \cite{kahya2023mcrood} propose a novel reconstruction-based OOD detector (MCROOD) for radar Range Doppler Images (RDIs), which accurately detects OOD human behaviors. \cite{Griebel21} propose a pointnet \citep{pointnet} based method which detects anomalous radar targets.

\subsection{Industrial 3D Scanner}

3D scanners play a crucial role in anomaly detection by capturing high-resolution, detailed point clouds of physical objects, enabling the identification of surface defects and structural deformations that may not be visible through traditional 2D imaging methods. For instance, the MVTec3D-AD dataset \citep{bergmann2021mvtec} employs the Zivid 3D camera, which offers a point precision of 0.11 mm, allowing for accurate surface inspection in industrial applications. Similarly, Real3D-AD \citep{liu2023real3d} utilizes a high-resolution binocular 3D scanner, the PMAX-S130, with a significantly higher point precision ranging from 0.011 mm to 0.015 mm.

High-resolution data alone is not enough for anomaly detection, as it also relies heavily on robust point cloud processing methods to accurately extract features. 
BTF \citep{btf} combines handcrafted 3D descriptors with the classic 2D PatchCore framework \citep{patchcore}, establishing a foundational approach to 3D anomaly detection. M3DM \citep{wang2023multimodal} builds on this by separately extracting features from point clouds and RGB images, which are then fused to improve detection performance. CPMF \citep{cpmf} enhances anomaly detection by rendering point clouds into 2D images from multiple viewpoints, extracting features using a pre-trained network, and integrating them for final prediction. Reg3D-AD \citep{liu2023real3d} introduces a registration-based approach that uses the RANSAC algorithm to align each input sample with a stored template before comparison during inference. IMRNet \citep{ashapenet} leverages PointMAE \citep{pang2022masked} to reconstruct clean, anomaly-free samples and identifies anomalies by comparing the reconstructed point cloud with the original input. More recently, diffusion models have been employed for reconstructing normal samples to facilitate anomaly detection \citep{zhou2024r3dad}.

\subsection{Magnetic Resonance Imaging}
Magnetic Resonance Imaging (MRI) is an imaging technique that uses magnetic fields and radio waves to generate 2D slices or 3D volumes of internal body organs \citep{miasurvey17}. For application such as unsupervised lesion detection, many existing works involve 2D convolutional neural networks, which treat MR images as RGB images \citep{BAUR2021101952,larsen2016autoencoding,baul18,weninger2023diffusion,ddpm2024,MANSOJIMENO202242,hybridcnn2020,faad21}. The advantage of 2D CNN is that some classic network structures \citep{he2015deep,vae13arxiv,aae15} can be reused and transfer learning is easy \citep{faad21}.

In some works \citep{kleesiek2016deep,kamnitsas2017efficient,milletari2016v}, 3D convolution is involved, which processes the entire MRI volume as a single input, maintaining spatial context across slices. 3D CNNs capture relationships between adjacent slices, making them more suitable for volumetric features or detecting anomalies that span across multiple planes, effective for anomalies that extend across slices.

\subsection{Muti-Sensor Fusion}
Sensor fusion combines data from multiple sensors to provide a more robust understanding of the environment. 

RGB-D camera captures both color (RGB) and depth (D) information from a scene, which benefits uncertainty estimation \citep{bauza2019omnipush}, road anomaly detection \citep{nguyen2022effective}, 3D symmetry detection with incomplete observation \citep{shi2022learning} and depth estimation in OOD situations \citep{kong2024robodepth}.

RGB-D data is quite useful for unsupervised anomaly detection. 
Unsupervised anomaly detection suffers from overgeneralization over neural networks, which leads to overlap between normal and abnormal distributions. Outlier Exposure \citep{oe} can be a useful strategy in this case. For example, Collaborative Discrepancy Optimization (CDO) \citep{cdo} enlarges the distribution margin by training the network against synthesized anomalies.  \cite{RudWeh2023} observe that the output of asymmetric teacher-student networks (AST) differs larger when encountering an anomaly, which can be easily adapted to multimodal data.

\cite{cpmf} propose Complementary Pseudo Multimodal Feature (CPMF), which aggregates the 3D feature and Multi-view 2D feature extracted by pretrained neural networks to summarize local and global geometry for anomaly detection.
\cite{wang2023multimodal} propose Multi-3D-Memory (M3DM), a multimodal model for anomaly detection and segmentation, composed of three main components. It converts 3D point group features into plane features through interpolation and projection, and performs image and point cloud fusion in an unsupervised manner. The final Decision Layer Fusion (DLF) module aggregates multimodal information from multiple memory banks and performs anomaly detection and segmentation. \cite{pmlr-v202-chu23b} propose Shape-Guided Dual-Memory (SGDM) learning, which is based on the synergy between two specialized expert models to improve 3D anomaly detection. One model focuses on 3D shape geometry to identify anomalies in the structure, while the other uses RGB information to detect appearance anomalies related to color. Specifically, SGDM employs Neural Implicit Functions (NIFs) to represent local shapes through signed distance fields, following the approach used in current 3D reconstruction methods \citep{nif}. \cite{zavrtanik2021draem} propose Depth-Aware Discrete Autoencoder (DADA), which learns a unified discrete latent space combining RGB and 3D data, tailored for 3D surface anomaly detection. \cite{zhou2024pointad} propose PointAD, a novel method leveraging CLIP's generalization ability for zero-shot 3D anomaly detection on unseen objects. It integrates 3D and 2D data through hybrid representation learning, optimizing text prompts using 3D points and 2D renderings. PointAD aligns point and pixel representations to detect and segment 3D anomalies effectively, while also incorporating RGB data for enhanced understanding.

\begin{figure*}
    \centering
    \begin{subfigure}{0.45\linewidth}
        \centering
        \includegraphics[width=\linewidth]{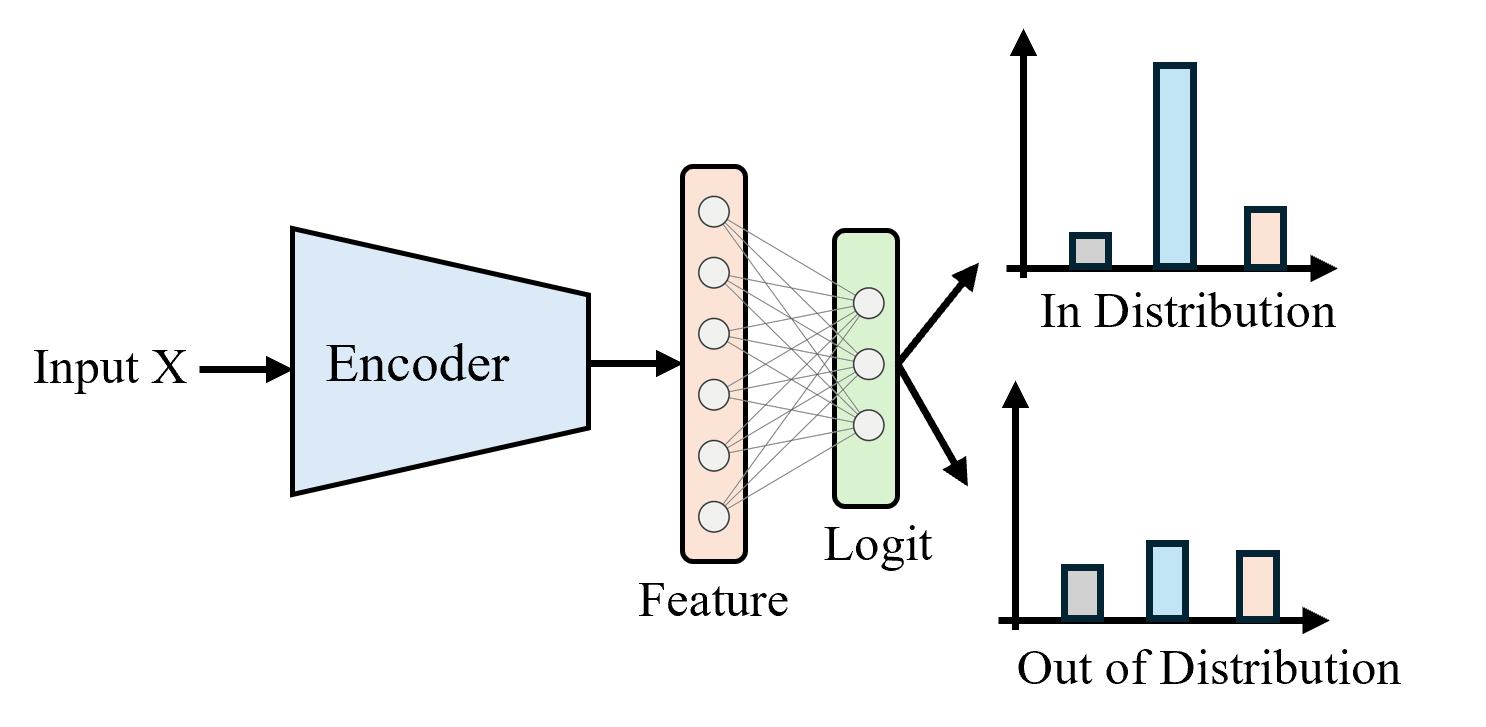}
        \caption{Logit-based Method}
        \label{logitbased}
    \end{subfigure}
    \begin{subfigure}{0.45\linewidth}
        \centering
        \includegraphics[width=\linewidth]{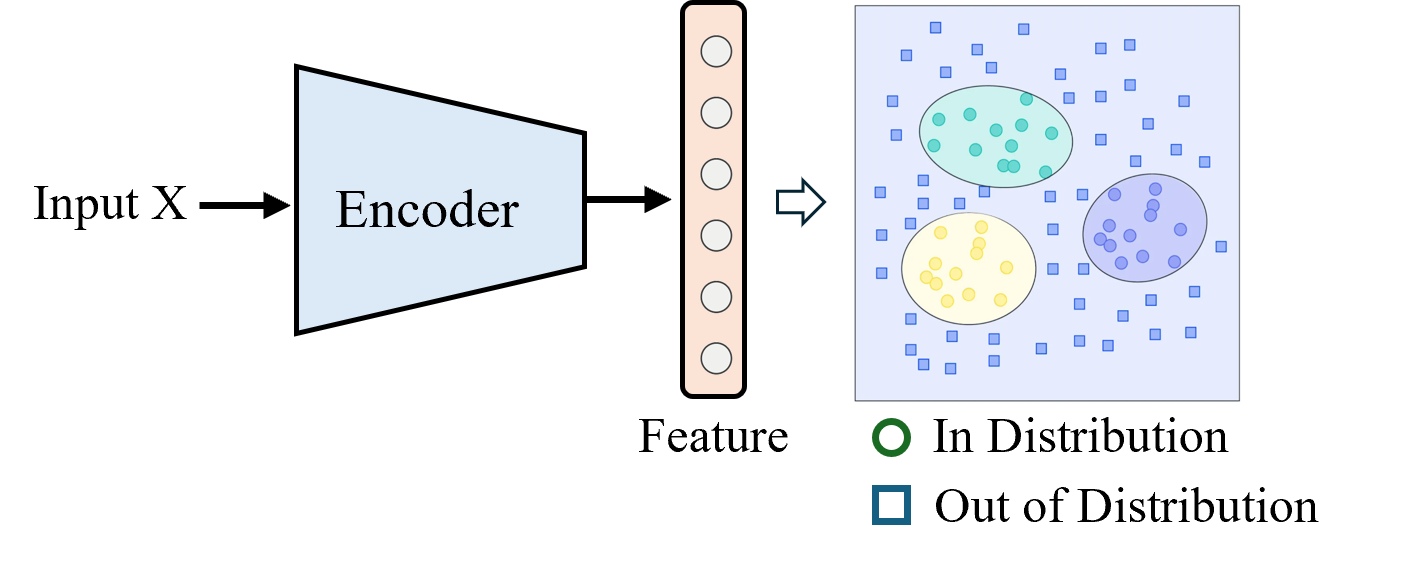}
        \caption{Feature-based Method}
        \label{featurebased}
    \end{subfigure}
    ~
    \begin{subfigure}{0.45\linewidth}
        \centering
        \includegraphics[width=\linewidth]{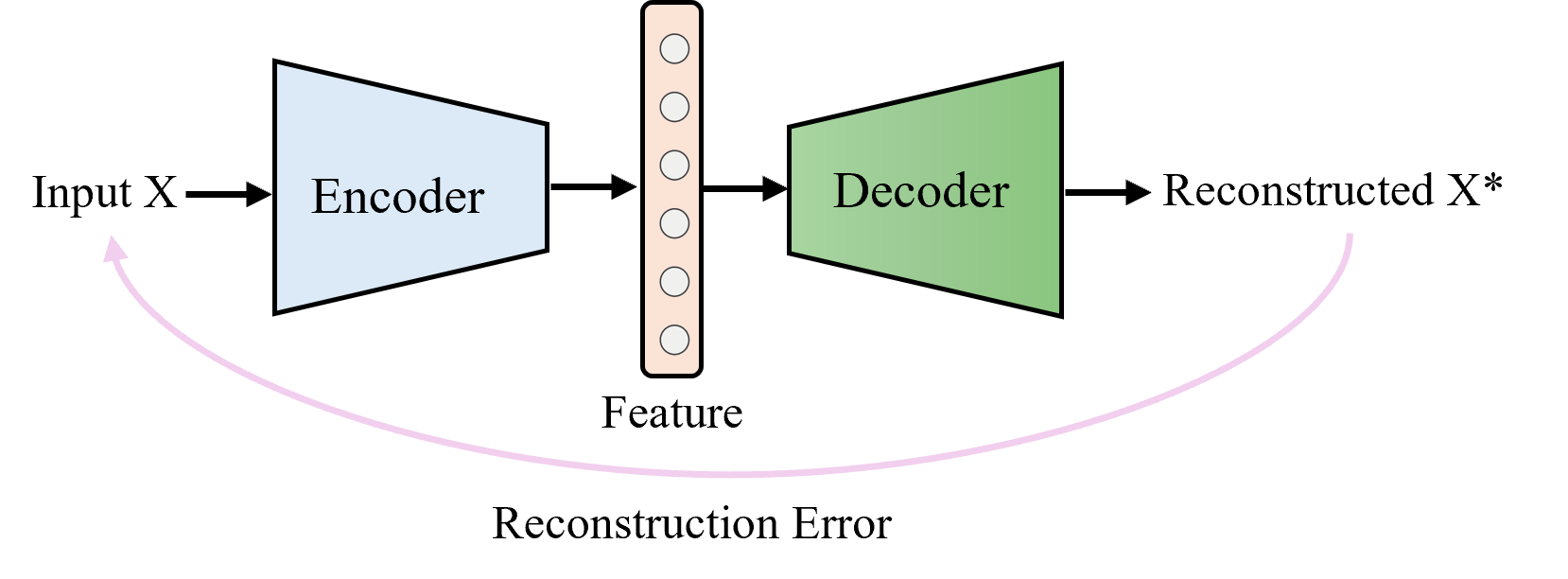}
        \caption{Reconstruction-based Method}
        \label{Reconstruction}
    \end{subfigure}
    \begin{subfigure}{0.45\linewidth}
        \centering
        \includegraphics[width=\linewidth]{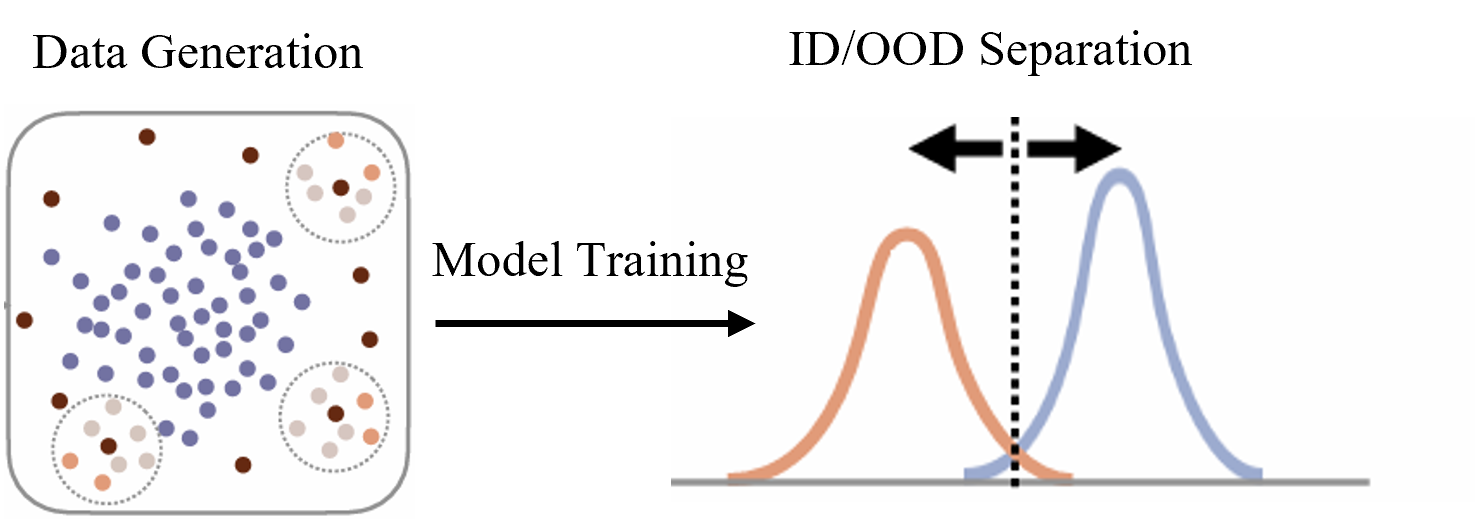}
        \caption{Generative Method}
        \label{TODO}
    \end{subfigure}
    \caption{Different types of OOD detection methods. (a) Logit-based OOD detection methods utilize the output of the final layer of deep neural networks. (b) Feature-based OOD detection methods typically measure OOD-ness based on statistical distance in the feature space (penultimate layer of DNN). (c) Meanwhile, reconstruction-based methods are self-supervised and measure OOD-ness based on the reconstruction error of input. (d) Generative methods improve the ID/OOD separation by synthesizing auxiliary OOD data.}
    \label{oodmethodsgrouped}
\end{figure*}

\section{Methods for OOD detection}
\label{sec:method}
Out-of-distribution (OOD) detection and anomaly detection methods may differ in specific details, but many share similar properties, allowing them to be categorized together. We group OOD detection methodologies into four categories, as summarized in \cref{oodmethodsgrouped}.

Logit-based OOD detection methods leverage the output of the final layer of deep neural networks. These methods can be easily applied to downstream tasks such as object detection \citep{itsc22,revisit2024} and semantic segmentation \citep{fishyscapes,smiyc} in a post-hoc manner.
Feature-based OOD detection methods evaluate OOD-ness based on statistical distances in the feature space (i.e., the penultimate layer of a deep neural network). While effective, achieving optimal performance often requires specialized training techniques \citep{csi,cider,PALM2024}.
Reconstruction-based methods adopt a self-supervised approach, measuring OOD-ness through the reconstruction error of the input.
Generative methods enhance the separation between in-distribution (ID) and OOD data by synthesizing auxiliary OOD samples.

Each of these groups offers distinct advantages and is suited to different use cases. We primarily focus on methods that can be adapted to 3D tasks in a post-hoc manner or implemented by simply modifying the backbone network.

\subsection{Logit-based OOD Detection}
Logit-based OOD detection methods capture OOD-ness based on the logit vector, the output of the last layer in a deep neural network. 
This type of method usually does not have special requirements for the neural network architecture and can be easily applied to 3D tasks, such as 3D object detection for autonomous driving \citep{itsc22,revisit2024}.

The most straightforward OOD detection method is Maximum Logit \citep{maxlogit}, which takes the maximum logit value as the OOD indicator. Maximum Logit is a strong baseline for OOD detection in image recognition and semantic segmentation.
After softmax normalization, the logit becomes a probability distribution with values between 0 and 1, which represents class probabilities. It has been shown that neural networks are more confident with in-distribution samples \citep{msp}, the softmax scores for in-distribution classes tend to be higher than those for OOD samples, making this a straightforward baseline. However, softmax struggles to generalize to different amounts of inputs \citep{veličković2024softmaxforsharpoutofdistribution}.
\cite{odin18iclr} propose ODIN, a method that applies small input perturbations to enlarge the softmax score gap between known and unknown objects. Generalized ODIN (G-ODIN)~\citep{godin20cvpr} builds upon ODIN~\citep{odin18iclr} by incorporating a specialized training objective called DeConf-C and selecting perturbation magnitude based on in-distribution data. Out-of-distribution input can result in a more uniform softmax distribution, which increases entropy \citep{agnostophobia18nips}. Energy-based OOD score \citep{energy} maps the logit to a single scalar, using a temperature-scaled LogSumExp (LSE). The variants \citep{du2022vos,pweal,rpl} are used in object detection and semantic segmentation. Later, JointEnergy \citep{jenergy} for multi-label classification. 

In addition, the energy-based OOD detector can also utilize auxiliary OOD data to better shape the energy gap between in-distribution and OOD samples \citep{energy}. This strategy of using auxiliary data (outlier) to increase ID/OOD differentiation is often known as Outlier Exposure \citep{oe}. 
Outliers can come from real-world datasets or data generation.  Early works promote uniform or high-entropy predictions for auxiliary OOD samples from other datasets~\citep{oe,agnostophobia18nips}. Later, OECC~\citep{oecc21neurocomputing} demonstrated that incorporating regularization for confidence calibration could further enhance the performance of OE. To make effective use of the typically large number of available OOD samples, some studies employ outlier mining~\citep{outliermining21ecml, ming2022posterior} or adversarial resampling~\citep{backgroundsampling20cvpr} techniques to generate a compact yet representative subset.
Outlier exposure is widely used in autonomous driving applications \citep{rpl,synboost,metaood,pweal,dense,mask2anomaly,fishyscapes,smiyc} and has shown strong performance.

Some methods also incorporate distance measures. OpenMax \citep{openmax} calculates the centre of each in-distribution class and builds a statistical model based on the distances of correctly classified samples. It leverages extreme value theory (EVT) to identify outliers by fitting a Weibull distribution to the tail of the distance distribution. \cite{cac} propose a straightforward method that trains the neural network to cluster in-distribution samples around the corresponding class anchor, and reject OOD samples based on softmin and Euclidean distance. 

The emergence of foundation models offers a fresh perspective on OOD logit-based OOD detection. \cite{mcm} introduce the Maximum Concept Matching (MCM) score, which estimates the OOD-ness of outputs from vision-language models like CLIP \citep{clip} by analyzing the maximum cosine similarity. The score can also be extended with negative labels~\citep{jiang2024negative}.  Based on the fact that vision-language pre-training aligns the feature spaces of text and image inputs, \cite{clipn} propose a prompt learning method that detects OOD samples in a zero-shot manner. Local regularized Context Optimization (LoCoOp) \citep{locoop} uses ID-irrelevant (such as background) of the image as the negative sample to perform prompt learning. \cite{negativeprompts} use a CoCoOP \citep{zhou2022cocoop} like strategy to learn negative prompts. Recently, \cite{npcvpr24} proposed a transferable negative prompt learning method to adapt to open vocabulary learning scenarios where new categories can be introduced.

\subsection{Feature-based OOD Detection}
Feature-based methods for OOD detection typically utilize feature embeddings from the penultimate layer of a deep neural network. \cite{md} employ the Mahalanobis distance (MD) to measure OOD-ness by calculating the distance between the feature vector of a test sample and the nearest class-conditional Gaussian distribution, where MD accounts for variable correlations and provides a robust metric for distinguishing OOD samples. \cite{ren2021simple} further refine this approach by introducing the relative Mahalanobis distance (RMD), which enhances detection performance, particularly for near-OOD samples.

To avoid assuming a specific distribution of the feature space, \cite{knn22} propose using the k-th nearest neighbor (KNN) distance between the test embedding and training embeddings as an OOD detection score. 
Another recent study \citep{she23} detects OOD-ness by matching the test sample's embedding with stored training embeddings using the inner product. \cite{vim} propose virtual logit matching, which applies principal component analysis (PCA) to the feature space of in-distribution data. The residual of the principal subspace is then used to construct a virtual logit as an OOD indicator. To improve computational efficiency, \cite{liu2024fast} suggest using the average feature distances from decision boundaries as an OOD score, reducing the complexity of feature-based methods.

Some methods utilize contrastive learning to form more compact clusters for known classes, thereby increasing the separation between in-distribution (ID) and OOD samples \citep{csi, knn22,sehwag2021ssd,Kim_2020_CVPR}.
\cite{csi} introduce the Contrastive Shifted Instance (CSI) approach, which contrasts in-distribution samples with auxiliary OOD samples generated through data augmentation. Studies \citep{csi,cider, PALM2024,li2024contrastive,mirzaei2025adversarially} also show that supervised contrastive learning can be effective without auxiliary OOD samples, as it promotes intra-class compactness and inter-class separation in the feature space. 

\begin{figure}[ht]
  \centering
  \includegraphics[width=\linewidth]{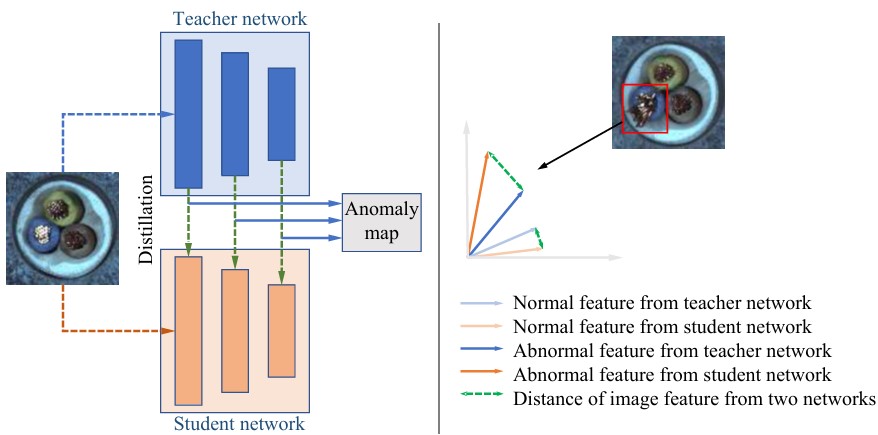}
  \caption{Overview of teacher-student models \citep{liu_industrial_survey}. During training, the teacher model teaches the student to extract features from normal samples. At inference, both networks produce similar features for normal images but differ for abnormal ones. By comparing their feature maps, we generate an anomaly score map, resize it to match the input image, and use it to identify anomalous regions and determine if the image is abnormal.
  }
  \label{fig:teacherstudent}
\end{figure}

Teacher-student models are frequently used in industrial anomaly detection \citep{Bergmann20,mkd21,wang2021student_teacher,RudWeh2023,yamada2021reconstruction,rd4ad,ikd22,Zhang_2024_CADAD} for both 2D and 3D inputs.
As shown in \cref{fig:teacherstudent}, during training, the teacher model teaches the student model to extract features of normal samples. During inference, normal images produce similar features in both networks, while abnormal images show distinct differences. Comparing the feature maps of the two networks generates anomaly score maps that help determine whether a test image is abnormal.
Although most of the related research focuses only on industrial anomaly detection, we believe that this network can also be transferred to other tasks.

In-distribution samples can also be noisy and polluted. \cite{mirzaei2024adversarially} propose Adversarially Robust Out-of-Distribution Detection (AROD), a method grounded in Lyapunov stability theory that guides both in-distribution and out-of-distribution samples toward distinct stable equilibrium states.

\subsection{Reconstruction-based OOD Detection}

The main concept behind reconstruction-based methods is that an encoder-decoder model trained on in-distribution (ID) data typically produces distinct results for ID and OOD samples. This variation in model performance can serve as a signal for identifying OOD samples.
A classic autoencoder consists of two components: an encoder that compresses the input data into a low-dimensional latent representation (often referred to as the latent space or bottleneck), and a decoder that reconstructs the input from this representation. The model is trained using in-distribution data to minimize a reconstruction loss, typically defined as the difference between the input and its reconstruction.
VAEs \citep{vae13arxiv} extend classic autoencoders by introducing a probabilistic framework. Instead of directly mapping input data into a fixed latent representation, VAEs model the latent space as a distribution, typically parameterized by a mean and variance. During training, the VAE optimizes two objectives: a reconstruction loss (similar to classic AEs) and a regularization term that forces the learned latent distribution to match a predefined prior distribution (e.g., a multivariate Gaussian). This regularization is achieved by minimizing the Kullback-Leibler (KL) divergence between the learned latent distribution and the prior. 

\begin{figure}[ht]
    \centering
    \includegraphics[width=0.95\linewidth]{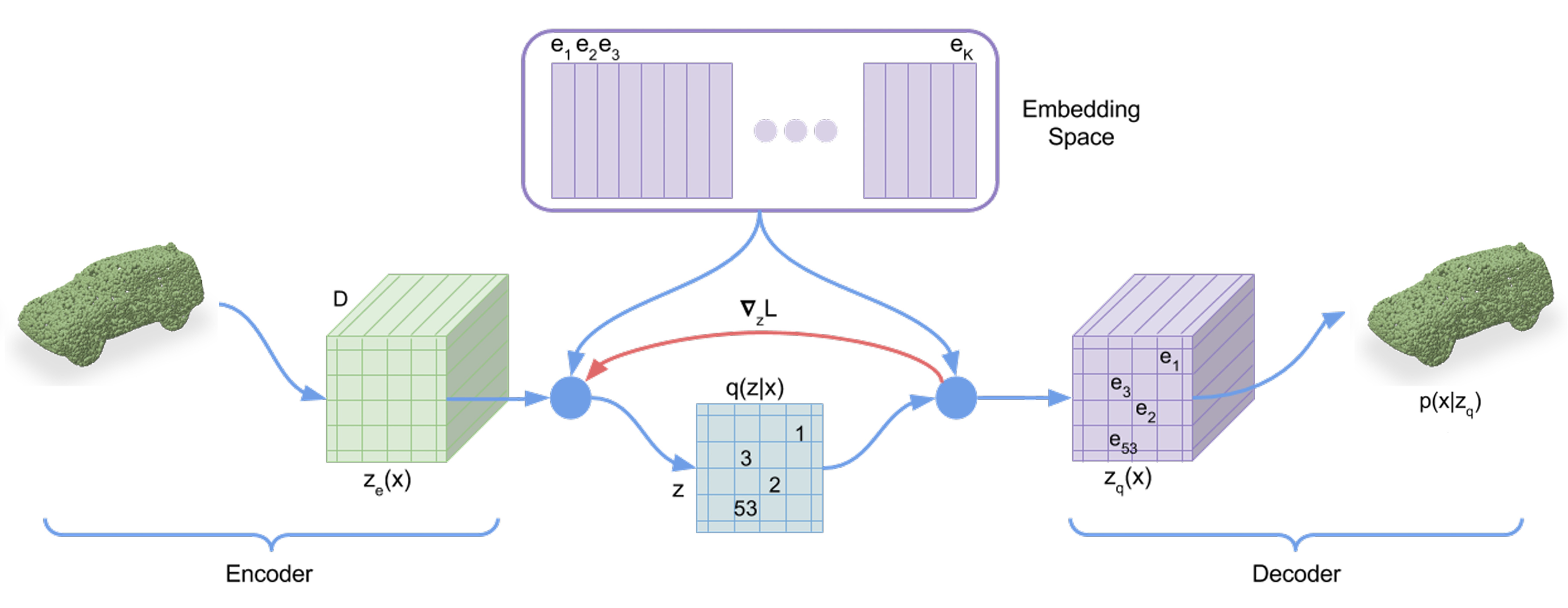}
    \caption{The architecture of Vector Quantised
Variational AutoEncoder (VQ-VAE) (modified after \citep{vqvae}). It learns a discrete latent space by quantizing the encoder output into a finite set of vectors (codebook), which helps the model avoid the problem of posterior collapse, a situation where the latent variables are effectively disregarded due to the dominance of a strong autoregressive decoder.}
    \label{fig:vqvae}
\end{figure}

Vector Quantised Variational AutoEncoder (VQ-VAE) \citep{vqvae} combines VAEs with vector quantization, enabling discrete latent representations.
As shown in \cref{fig:vqvae}, VQ-VAE learns a discrete latent space by quantizing encoder outputs into a finite set of codebook vectors. This approach mitigates posterior collapse, a common issue in standard VAEs \citep{vae13arxiv} where latent variables are ignored due to the dominance of a powerful autoregressive decoder, enabling efficient generation of high-fidelity, diverse images and shapes \citep{vqvae,3dqd}.
Discrete representations are suitable for various modalities, not only limited to images but also applicable to time series anomaly detection and 3D shape generation \citep{3dqd,chen2024sar3d,talukder2024totem}. In a recent work \citep{talukder2024totem}, VQ-VAE has been used as a generalized method for time series anomaly detection across domains.

As another variant of VAEs, Adversarial Autoencoder (AAE) \citep{aae15} replaces the KL divergence regularization with an adversarial network. The adversarial network acts as a proxy to align the learned latent distribution with the prior. Unlike the KL divergence, adversarial training does not favor specific modes in the distribution and is fully differentiable, making it a more flexible approach. 

GMVAEs \citep{gmvae20aaai} extend VAEs by replacing the unimodal prior with a Gaussian mixture model. This increases the expressive power of the latent space, allowing the model to capture more complex distributions. 

Memory-augmented autoencoders (MemAE) \citep{Gong_2019_ICCV} enhance traditional autoencoders by incorporating a memory module to better detect anomalies. The memory module stores prototypical patterns of normal data, ensuring that reconstructions closely align with previously seen normal patterns. This approach makes it easier to identify anomalies based on higher reconstruction errors.

Reconstruction-based methods are also widely used in open-set recognition (OSR), where maintaining in-distribution classification performance is essential. To achieve this, C2AE \citep{c2ae19cvpr} employs a fixed visual encoder derived from standard multi-class training and trains a decoder conditioned on label vectors, and uses extreme value theory (EVT) to model reconstruction error.
Subsequent approaches introduce conditional Gaussian distributions, encouraging latent features to align with class-specific Gaussian models. This strategy facilitates both the classification of known samples and the rejection of unknown ones \citep{cgdl20cvpr}. 
\cite{gen} further combine an auto-encoder with a multi-task classifier, optimizing both a self-supervision loss and a classification loss. The self-supervised learning component applies random transformations to improve the quality of the learned features.
Similarly, \cite{crl} propose Deep hierarchical 
reconstruction net, which enhances feature representation by integrating classification and input reconstruction during model training.

\subsection{Generative Models for OOD Detection}
 
A major challenge for OOD detection is the absence of supervisory signals from unknown data, leading models to make overly confident predictions on OOD samples \citep{du2022vos}.
Outlier exposure \citep{oe} heavily relies on the assumption that OOD training data is readily available, which may not always be practical. In real-world scenarios, OOD data can be rare, domain-specific, or expensive to collect, limiting the effectiveness of these approaches. Consequently, there is a need for techniques that can generate synthetic OOD samples to achieve ID/OOD separation without requiring access to extensive OOD datasets. Data generation not only provides a scalable solution to this challenge.

Existing approaches have explored various strategies for OOD sample generation. GANs are frequently used to create synthetic OOD samples, enforcing uniform model predictions \citep{confcal18iclr}, restore in-distribution samples \citep{Schlegl_2017,SCHLEGL201930}, generating boundary samples in low-density regions \citep{oodsg19nipsw}, estimate class activations \citep{gopenmax} or producing high-confidence OOD samples \citep{confgan18nipsw,auroc}. Additionally, meta-learning techniques have been utilized to enhance sample generation \citep{maml20nips}.

Recent works such as Virtual Outlier Synthesis (VOS) \citep{du2022vos} focus on synthesizing virtual outliers in the low-likelihood regions of feature space, leveraging its lower dimensionality for more efficient generation. VOS employs a parametric approach, modeling the feature space as a class-conditional Gaussian distribution, while NPOS \citep{npos2023iclr} uses a non-parametric strategy to generate outlier ID data. Recognizing that synthetic OOD data may sometimes be irrelevant, Distributional-agnostic Outlier Exposure (DOE) \citep{wang2023out} generates challenging OOD samples to train detectors using a min-max learning framework, and Auxiliary Task-based OOD Learning (ATOL) \citep{zheng2023out} incorporates auxiliary tasks to mitigate errors during OOD generation.

With the advancement of point cloud and generation techniques \citep{pointflow,Luo_2021,XIANG2024105207,kirby2024logen,kang2025multiview,lookoutside,gao2024catd}, these OOD detection methods have become increasingly applicable to 3D vision tasks.



\section{Evaluation Metrics of OOD Detection Performance}
In out-of-distribution detection (OOD) benchmarks, test samples are labeled as either in-distribution or out-of-distribution. The OOD detector assigns a confidence score to each sample, indicating how likely the model considers it to be normal. Samples with confidence scores below a predefined threshold are classified as OOD.

\subsection{Classification Metrics}
OOD detection in object-recognition tasks or semantic segmentation can be framed as a binary classification problem, where each input is assigned an ID or OOD label. Consequently, standard classification metrics are often used to evaluate the performance of OOD detection.

\begin{figure}[ht]
  \centering\includegraphics[width=\linewidth]{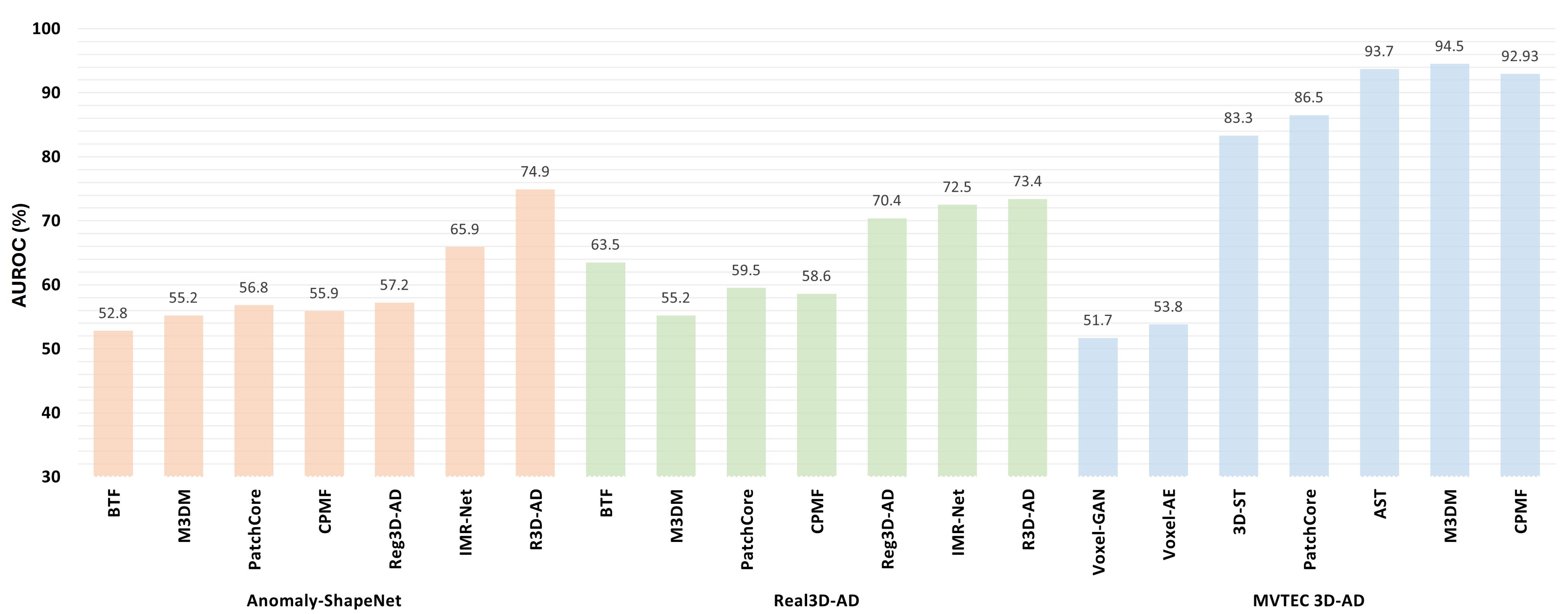}
  \caption{Comparison of AUROC performance across different methodologies on three commonly used benchmarks: Anomaly-ShapeNet~\citep{ashapenet}, Real3D-AD~\citep{liu2023real3d}, and MVTec 3D-AD~\citep{bergmann2021mvtec}.}
  \label{result}
\end{figure}

\noindent\textbf{AUROC}
The area under the receiver operating characteristic curve (AUROC) is a widely used metric in 3D OOD detection \citep{revisit2024,oodnogpu,itsc22}.
It is calculated by treating OOD samples as positive cases and normal samples as negative \citep{msp}, generating a range of true positive rates (TPR) and false positive rates (FPR) at different thresholds.
\cref{result} shows a comparison of AUROC performance across three benchmarks: Anomaly-ShapeNet~\citep{ashapenet}, Real3D-AD~\citep{liu2023real3d}, and MVTec 3D-AD~\citep{bergmann2021mvtec}. 
In general, recent models such as M3DM~\citep{wang2023multimodal} and R3D-AD~\citep{zhou2024r3dad} demonstrate superior performance, largely due to innovations in detection pipeline and training process.
Moreover, the overall trend indicates that MVTec 3D-AD consistently outperforms the other two benchmarks. This is likely attributable to its incorporation of both RGB and 3D data, which enables the use of pretrained 2D models. In contrast, Anomaly-ShapeNet and Real3D-AD show more modest and variable results across methods. These findings highlight the advantages of leveraging richer data modalities and pretrained models to improve 3D anomaly detection performance.

\noindent\textbf{AUPR}
Similarly, precision and recall can be used to compute the area under the precision-recall curve (AUPR). AUPR can be calculated in two ways: either treating ID (AUPR-S) or OOD (AUPR-E) as the positive class \citep{revisit2024}. AUPR is particularly useful when classes are imbalanced \citep{revisit2024,smiyc,fishyscapes}. In both AUROC and AUPR, higher values indicate better detection performance. AUPR can be calculated at either pixel level \citep{fishyscapes,smiyc} or instance level \citep{Zhang_2023_CVPR}.

\noindent\textbf{FPR @ 95 TPR}
(False Positive Rate at 95\% True Positive Rate) indicates how many normal samples are being incorrectly labeled as OOD when the model achieves a 95\% detection rate for actual OOD samples \citep{itsc22,revisit2024}. A lower FPR @ 95 TPR indicates better model performance since fewer normal samples are mistakenly identified as OOD.

\subsection{Object Detection Metrics}

\noindent\textbf{mAP}
Mean Average Precision (mAP) is commonly used in object detection tasks to evaluate how well a model detects objects across multiple classes \citep{Liu2018DeepLF}.  \cite{oodnogpu} consider OOD samples as one class and calculate the mAP across all classes.  In this way, undetected OOD samples are also taken into account. However, grouping all OOD samples into a single class overlooks their diversity.

\noindent\textbf{U-Recall}
Object detectors with a closed-set assumption tend to ignore unknown objects as background \citep{elephant}. U-recall (unknown recall) indicates how well the bounding boxes of OOD samples are detected. This metric is widely used in open-set/open-world object detection tasks \citep{owdetr,randbox,orth24,li2024open,liu2024yolouniow}, but has not been widely adopted by 3D OOD benchmarks.

\subsection{3D Anomaly Detection}
For 3D anomaly detection, classification metrics like AUROC and AUPR are well-suited for identifying anomalous instances. However, when the task involves localizing anomalies, an additional specialized metric is necessary.

\noindent\textbf{AU-PRO}
The AU-PRO (Area Under the Per-Region Overlap) \citep{bergmann2021mvtec} measures the accuracy of unsupervised anomaly localization by evaluating how well predicted anomalies overlap with the ground truth. It calculates the average overlap across all ground truth components at different thresholds, and then plots these values against false positive rates. The final AU-PRO score is obtained by integrating the curve up to a limited false positive rate and normalizing the area to [0, 1].

\section{Distribution Distance Taxonomy}
\begin{table}[ht]
    \centering
    \caption{Distance taxonomy for OOD detection.}
    \begin{tabular}{ll}
        \hline
        \textbf{Distance Metric} & \textbf{Formula} \\ 
        \toprule
        Euclidean Distance & \rule{0pt}{18pt} $D_E(z) = \sqrt{\sum_{i=1}^n (z_i - c_i)^2}$ \\ 

        Mahalanobis Distance  & \rule{0pt}{18pt} $D_M(z) = \sqrt{(z - \mu)^T \Sigma^{-1} (z - \mu)}$ \\
        Inner Product  &  \rule{0pt}{18pt} $<z, c> = z^Tc$ \\
        Cosine Similarity & \rule{0pt}{18pt} $\cos<z, c> = \frac{z^Tc}{\|z\| \|c\|}$ \\
        KL Divergence & $\min_k, D_{KL}[p(y\mid x) \|, d_k]$ \\ \bottomrule
    \end{tabular}
    \label{tab:distance_metrics}
\end{table}


OOD detection focuses on identifying when data comes from a different distribution than the one used during training, often referred to as a distributional shift \citep{yang2024generalized}. Distribution distance offers a practical way to quantify this shift. The intuition is that OOD samples are expected to be farther from the centroids or prototypes of in-distribution (ID) classes. By calculating the distances between an input sample and class-specific prototypes \citep{cac,cider,PALM2024,she23,openmax}, or the overall ID distribution \citep{md}, we can estimate how likely the sample belongs to the ID data. As shown in \cref{tab:distance_metrics}, we summarize commonly used distance-based metrics for OOD detection. These metrics provide a generalizable approach that can be applied in both feature space and logit space, making them broadly applicable across different OOD detection methods.

\textbf{Euclidean distance} is a simple tool for OOD detection \citep{fss20arxiv}. It measures the straight-line distance between two points, such as an input sample and the centroid of in-distribution (ID) classes. OOD samples are expected to have larger distances from these centroids than ID samples. The Euclidean distance is computed as:
\begin{align}    
\text{Euclidean Distance:} \quad D_E(z) = \sqrt{\sum_{i=1}^n (z_i - c_i)^2}, 
\end{align}

where $z$ is the input sample, $c$ is the class prototype, and $n$ is the feature dimension. We can classify samples as ID or OOD by comparing $D_E(z)$ to a threshold.

\textbf{Mahalanobis distance} is often preferred to Euclidean distance for OOD detection because it considers the input feature to the distribution distance \citep{md}.  Mahalanobis incorporates the covariance of the in-distribution (ID) data. It is defined as:
\begin{align}    
\text{Mahalanobis Distance:} & \quad D_M(z) = \sqrt{(z - \mu)^T \Sigma^{-1} (z - \mu)},
\end{align}

where $z$ is the input sample, $\mu$ is the feature mean of class prototypes, and \(\Sigma\) is the covariance matrix of the ID data. \cite{ren2021simple} further refine this approach by introducing the relative Mahalanobis Distance (RMD), and computing the ratio of Mahalanobis distances between the two regions of pixels.
  
In high-dimensional space, Euclidean distances can become ambiguous due to the "curse of dimensionality," where all points tend to appear equidistant \citep{towardosr13pami}.

\textbf{Cosine similarity} becomes a useful OOD detector. It measures the angular similarity between vectors, focusing on their direction alignment rather than magnitude \citep{cosinesim20accv,svae20eccv,knn22,cider}.

The cosine similarity is defined as:
\begin{align}    
 \quad \cos<z, c> = \frac{z^T c}{\|z\| \|c\|},
\end{align}    
where $z$ is the input vector and \(c\) is the class prototype. Higher similarity indicates closer alignment to the in-distribution data, making it effective for distinguishing OOD samples.

Without normalization, the inner product of the feature vector and the class prototype can also serve as an OOD score, which is theoretically defined as an approximation of the modern Hopfield energy \citep{she23}.

\textbf{Kullback-Leibler (KL) divergence} can also be used to measure how much the predicted probability distribution of a model for an input sample differs from the expected in-distribution (ID) data. For example, \cite{maxlogit} propose KL Matching, which compares the softmax posterior distribution of the network with class prototypes to compute an OOD score. Specifically, 
$k$ distributions $d_k$ are computed, one for each class. For a new test input $x$, the score will be $\min_k, D_{KL}[p(y\mid x) \|, d_k]$.

\begin{figure}[thbp]
  \centering
  \includegraphics[width=\textwidth]{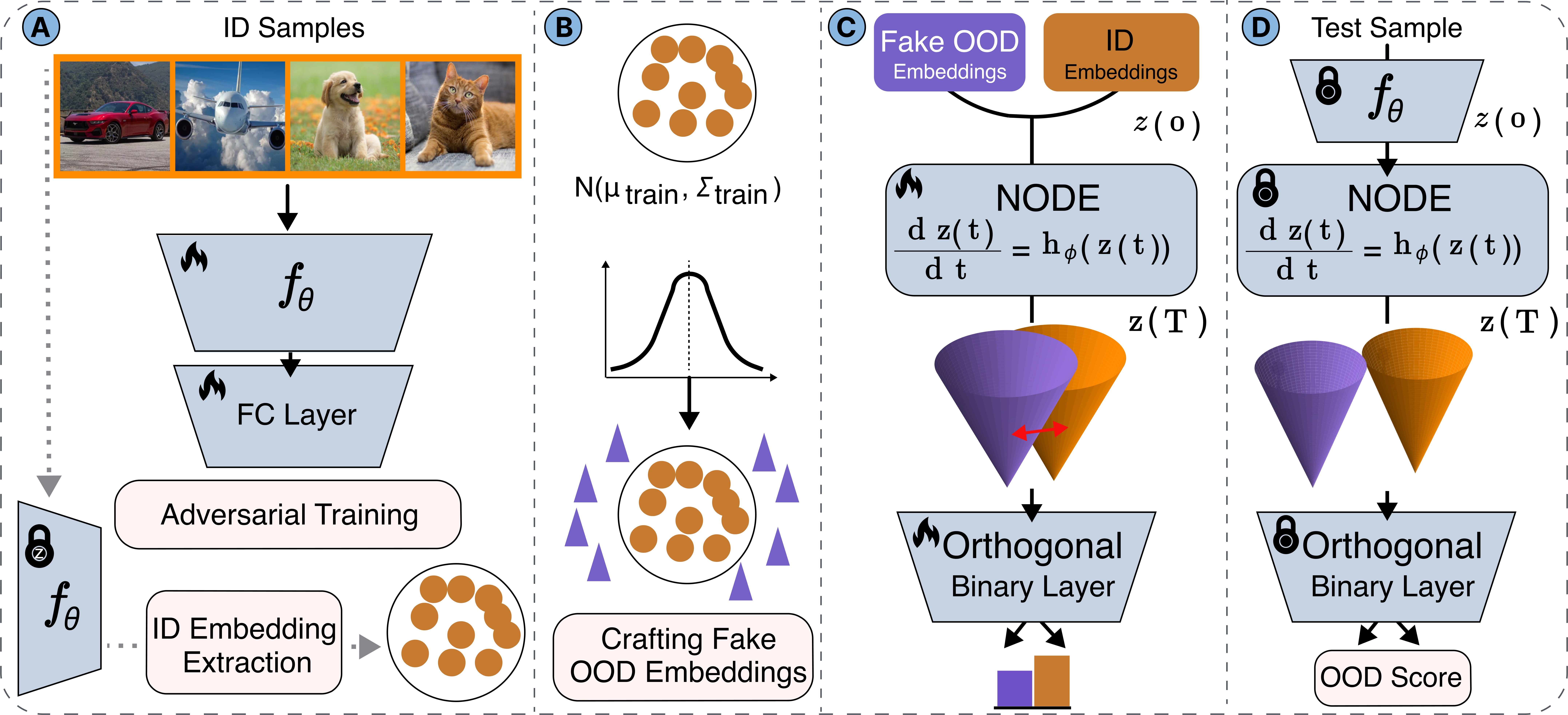}
  \caption{
The architecture of Lyapunov OOD detection \citep{mirzaei2024adversarially}. The model enhances OOD detection by: (A) performing adversarial training on a classifier using only ID samples, (B) generating fake OOD embeddings to create balanced ID and OOD classes, (C) incorporating a NODE layer and an Orthogonal Binary Layer to stabilize the system dynamics and (D) calculate the OOD score using the probability output from the Orthogonal Binary Layer.}
  \label{fig:lyapunov}
\end{figure}

\textbf{OOD robustness to adversary attack.} The performance gap between OOD detectors on clean versus adversarially perturbed data poses significant safety concerns in critical applications like medical diagnosis and autonomous driving. To enhance robustness against adversarial perturbations, Neural Ordinary Differential Equations (NODEs) incorporate the Lyapunov stability theorem \citep{mirzaei2024adversarially}. This approach models ID and OOD data as distinct stable equilibrium points, visualized as colored cones in \cref{fig:lyapunov}, where small perturbations decay over time.
To determine the ID-OOD boundary, we estimate a ($K$)-class-conditional Gaussian distribution. Fake OOD embeddings ($r$) are sampled from the feature space of class $(j) ((j = 1, \dots, K))$ when the probability falls below threshold ($\beta$):
\begin{align}
\frac{1}{(2\pi)^{d/2}|\Sigma_j|^{1/2}} \exp\left(-\frac{1}{2}(r - \mu_j)^T \Sigma_j^{-1} (r - \mu_j)\right) < \beta.
\label{eq:r-class}
\end{align}
The mean vector ($\mu_j$) and covariance matrix ($\Sigma_j$) of the (j)-th class training samples in feature space are computed as:
\begin{align}
\mu_j &= \frac{1}{n_j} \sum_{i:y_i=j} f_{\theta}(x_i), \\
\Sigma_j& = \frac{1}{n_j-1} \sum_{i:y_i=j} (f_{\theta}(x_i) - \mu_j)(f_{\theta}(x_i) - \mu_j)^T,
\end{align}
where ($f_{\theta}$) represents the encoder extracting ID embeddings from training samples ($x$).
\section{Open Challenges and Future Research Directions}
OOD detection extends beyond conventional settings and can be examined in a broader context. Failure detection, for instance, encompasses misclassification, covariate shifts (including corruption and domain shifts), and novel class shifts (OOD) \citep{jaeger2023a}, yet research in this area remains relatively limited. \cite{devries2018learning} suggests that estimating neural network uncertainty can aid in OOD detection, while \cite{sirc} introduce Softmax Information Retaining Combination (SIRC), a unified confidence score for detecting both misclassified known samples and OOD instances. Achieving trustworthy machine learning requires a holistic uncertainty estimation framework that accounts for diverse data modalities, domains, and applications, making this a critical direction for future research.

OOD detection in 3D extends beyond traditional image-based approaches, introducing new challenges in handling geometric variations, sensor noise, varying density, and spatial transformations \citep{3dptsurvey}. In addition, most OOD studies \citep{itsc22,li2024open,liu2024yolouniow,revisit2024} focus on static datasets, ignoring temporal variations inherent in real-world applications like autonomous driving and robotics.
Future research should investigate the role of sequential dependencies and motion patterns in OOD detection, particularly in dynamic settings. Spatiotemporal models should accurately identify OOD instances in video sequences or multi-frame point cloud data is crucial for enhancing real-world applicability. 
Establishing comprehensive benchmarking frameworks with standardized evaluation metrics across dynamic and multimodal datasets is another promising direction, as it would facilitate systematic comparisons and progress in OOD detection.

Balancing OOD generalization and detection accuracy is also crucial yet challenging, as improving one often compromises the other \citep{wang2024bridging}. To address this, future research should focus on adaptive thresholding mechanisms that dynamically adjust detection sensitivity based on environmental context, reducing false positives while maintaining high recall. Contrastive learning can enhance feature separability, helping distinguish unknown instances more effectively. Additionally, self-supervised learning offers a promising direction by enabling models to generalize without explicit OOD labels. Multi-objective optimization, combining regularization, ensembling, and active learning, could further refine the trade-off, ensuring robust and reliable 3D OOD detection in real-world applications.

3D OOD detection can also intersect with other fields, such as open world recognition \citep{openmax}, which not only identifies OOD objects but also adaptively incorporates novel object categories over time.  Future work could focus on developing unified models that leverage both geometric and semantic information, allowing them to discern subtle differences in complex 3D structures while continuously updating their class representations. 

Moreover, Vision-Language Models (VLMs) \citep{clip, liu2023llava, Qwen-VL, Qwen2-VL, xu2024pointllm, li2022blip} demonstrate exceptional generalization capabilities by learning rich multimodal representations from large-scale datasets. Their application to 3D OOD detection presents numerous research opportunities. Recent studies have leveraged VLMs to align 3D representations, facilitating open-vocabulary learning on 3D data \citep{pointclip22, Huang2022CLIP2PointTC, uclip23, lidarclip, lu2023open, zhu2023object2scene} and enabling models to generalize beyond their training distribution. Despite these advancements, little research has investigated whether VLMs can reliably detect OOD instances in 3D spaces. Integrating open-vocabulary learning with OOD detection in 3D data could pave the way for zero-shot or transferable OOD detection \citep{mcm, jiang2024negative, npcvpr24}, allowing models to identify novel objects without requiring explicit supervision.

In addition, for generalized 3D OOD detection, particularly in anomaly detection for complex shapes, adopting more expressive geometric representations and leveraging negative auxiliary data generation could improve robustness against irregular shape anomalies while ensuring consistent generalization across different object classes.


\section{Conclusion}

This paper explores 3D OOD detection from the perspectives of downstream applications and sensor modalities while also summarizing general OOD detection methodologies applicable to 3D settings. Our objective is to provide readers with a clear understanding of mainstream approaches, facilitate the selection of appropriate baselines for 3D OOD detection, and inspire the development of future solutions. We aim to inspire stronger connections between theoretical OOD research and real-world 3D applications by outlining key insights, challenges, and research directions.

\section*{ACKNOWLEDGMENT}

The authors acknowledge the financial support from the University of Melbourne through the Melbourne Research Scholarship.

\bibliographystyle{apalike} 
\bibliography{bibfull}

\end{document}